\documentclass{article}
\usepackage{iclr2027_conference,times}
\usepackage[utf8]{inputenc}
\usepackage[T1]{fontenc}
\usepackage{amsmath,amssymb,amsfonts,mathtools}
\usepackage{graphicx,booktabs,array,multirow,adjustbox}
\usepackage[table]{xcolor}
\definecolor{phoenixgray}{gray}{0.93}
\newcommand{\posgain}[1]{\textcolor{red}{#1}}
\newcommand{\neggain}[1]{\textcolor{blue}{#1}}

\usepackage{algorithm,algorithmic}
\usepackage{xspace}
\usepackage{microtype}
\usepackage{url}
\usepackage{wrapfig}
\usepackage{float}
\usepackage[hidelinks]{hyperref}
\usepackage{placeins}

\hypersetup{pdftitle={PhoenixSR: Generative Heterogeneous Distillation Unleashes Efficient Models for Real-World Super-Resolution},pdfauthor={Xin Di, Mingyu Shi, Yuanfei Bao, Long Peng, Yue Zhao, Jiaming Guo, Renjing Pei, Xueyang Fu, Yang Cao, Zheng-Jun Zha}}
\title{PhoenixSR: Generative Heterogeneous Distillation Unleashes Efficient Models for Real-World Super-Resolution}
\author{
\textbf{Xin Di}$^{1,*}$ \quad \textbf{Mingyu Shi}$^{1,*}$ \quad
\textbf{Yuanfei Bao}$^{1}$ \quad \textbf{Long Peng}$^{1,\dagger}$ \quad
\textbf{Yue Zhao}$^{2}$ \quad \textbf{Jiaming Guo}$^{2}$ \\[-1pt]
\textbf{Renjing Pei}$^{2}$ \quad \textbf{Xueyang Fu}$^{1,\dagger}$ \quad
\textbf{Yang Cao}$^{1,\dagger}$ \quad \textbf{Zheng-Jun Zha}$^{1}$ \\[-1pt]
{\normalfont $^{1}$University of Science and Technology of China} \quad
{\normalfont $^{2}$Huawei Technologies Ltd.} \\[-1pt]
{\normalfont\scriptsize $^{*}$Equal contribution. \quad $^{\dagger}$Corresponding authors.} \\[-1pt]
{\normalfont\scriptsize
\texttt{\{dx9826,shimy2003,longp2001\}@mail.ustc.edu.cn} \quad
\texttt{\{xyfu,forrest\}@ustc.edu.cn}}
}
\iclrfinalcopy

\begin{document}
\maketitle
\lhead{Preprint}
\vspace{-10mm}
\begin{figure}[H]
    \centering
    \includegraphics[width=\textwidth]{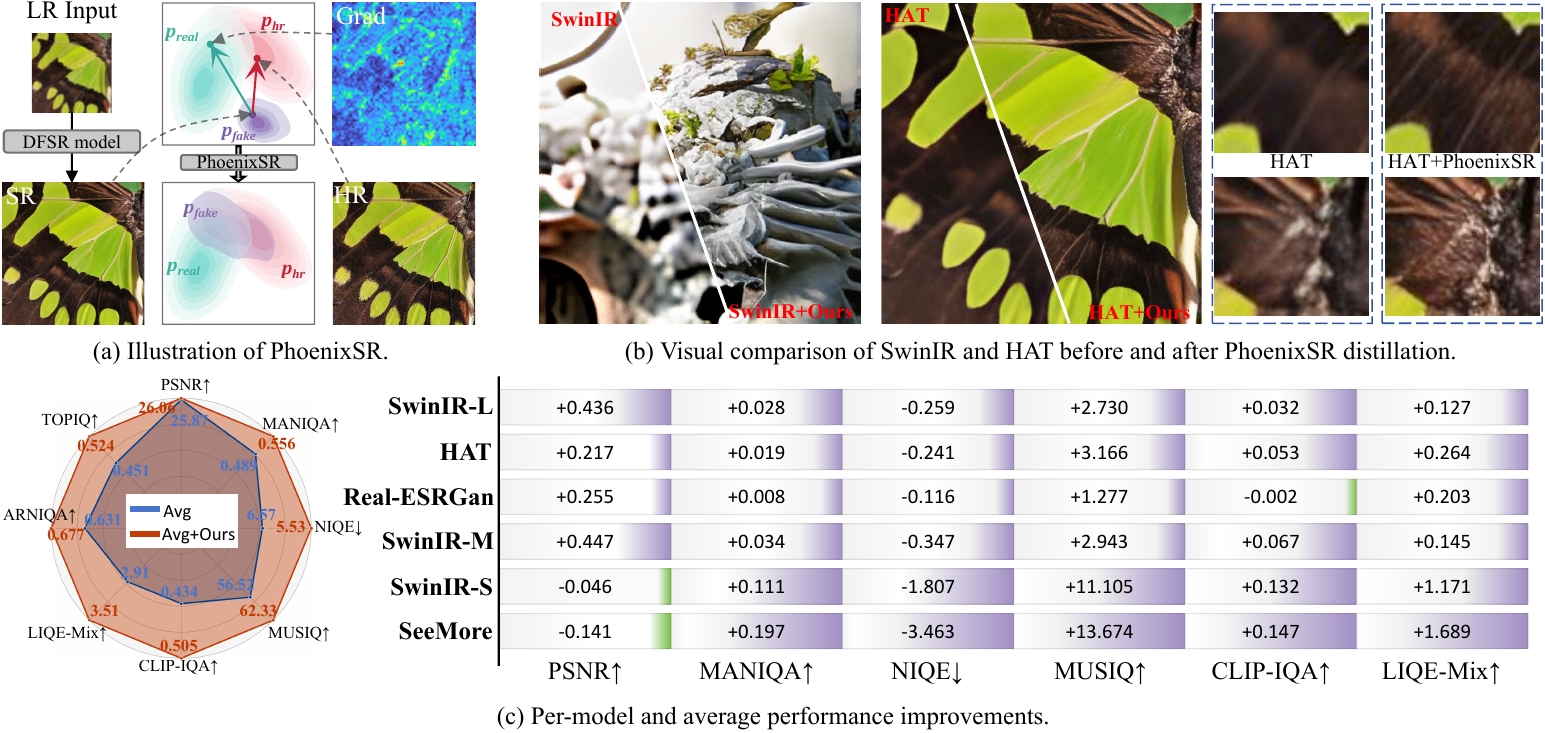}
    \caption{
    \textbf{Overview and performance of PhoenixSR.}
    (a) PhoenixSR transfers diffusion generative priors to diffusion-free SR students through generative heterogeneous distillation.
    (b) Visual comparisons of SwinIR and HAT before and after PhoenixSR distillation.
    (c) Average and model-wise performance improvements brought by PhoenixSR on RealSR.
    }
    \label{fig:first_figure}
\end{figure}
\vspace{-5mm}

\begin{abstract}
\vspace{-6mm}
Real-world image super-resolution (SR) requires recovering perceptually realistic high-resolution images from complex low-resolution observations while preserving faithful content. Diffusion-based SR benefits from strong generative priors but incurs substantial computational overhead, whereas feed-forward CNN and Transformer SR models are efficient yet often struggle to recover realistic high-frequency details. This motivates a natural question: \textit{can diffusion priors be transferred to existing diffusion-free SR networks without introducing diffusion components at inference time?} To this end, we propose \textbf{PhoenixSR}, a generative heterogeneous distillation framework that transfers diffusion priors to independently designed feed-forward SR networks through score-based distribution matching. Rather than aligning heterogeneous features or imitating sampled diffusion outputs, PhoenixSR uses the pretrained diffusion model as distribution-level supervision, while paired SR supervision preserves reconstruction fidelity. To make distribution matching effective for fidelity-sensitive SR, we introduce \textbf{Heterogeneous Distribution Adaptation}, which adapts the target score to the SR domain, improves tracking of the evolving student distribution, and anchors training with paired supervision. We further employ \textbf{Directional Reliability Weighting}, a lightweight residual-consistency-based reweighting strategy that reduces unstable distributional guidance. All diffusion-related components are removed after training, leaving the original student architecture and inference cost unchanged. Experiments on three SR benchmarks and six feed-forward backbones, including SwinIR, HAT, Real-ESRGAN, and SeeMoRe, show consistent perceptual improvements with largely preserved reconstruction fidelity.
\end{abstract}

\section{Introduction}

Single image super-resolution (SISR) aims to reconstruct a high-resolution (HR) image from its low-resolution (LR) observation and has been extensively studied in computer vision~\citep{dong2014learning,dong2016image,kim2016accurate,lim2017enhanced,zhang2018rcan,dai2019san,zhang2020rnan,niu2020han,liang2021swinir,chen2023hat}. While early methods mainly consider synthetic degradations, real-world images often suffer from complex and unknown degradation processes, including blur, sensor noise, compression, and camera processing. Real-world SR is therefore particularly challenging, as the restored image should be perceptually realistic while remaining faithful to the input content~\citep{cai2019toward,zhang2021designing,wang2021realesrgan,yi2025fine,pengtowards,peng2024efficient,peng2024unveiling}.

Recent diffusion-based super-resolution (DBSR) methods achieve impressive perceptual quality by exploiting generative priors learned from large-scale natural image distributions~\citep{saharia2022sr3,rombach2022high,wang2023stablesr,yue2024resshift,lin2024diffbir,wu2024seesr,yang2024pixel,yu2024scaling}. However, iterative diffusion sampling requires repeated network evaluations and introduces substantial computation and latency~\citep{ho2020denoising,song2021ddim,nichol2021improved,lu2022dpm}. Although recent methods reduce the number of sampling steps~\citep{wang2024sinsr,wu2024osediff,chen2025adcsr,yi2026gdpo,deng2026op4ksr,deng2026joint}, their inference models still rely on diffusion backbones or diffusion-derived generation pipelines. In contrast, diffusion-free SR (DFSR) models directly reconstruct HR images through a single feed-forward pass~\citep{dong2014learning,kim2016accurate,lim2017enhanced,zhang2018rcan,liang2021swinir,chen2023hat,kong2022rlfn,li2020lapar,sun2023safmn,deng2026ihmambasr,peng2025pixel}. They are substantially more efficient, but often lack strong generative priors and tend to produce over-smoothed details under severe degradations. This raises a fundamental question: \emph{can the generative prior of a pretrained diffusion model be transferred to existing diffusion-free SR networks without introducing diffusion components at inference time?}

Knowledge distillation provides a natural way to transfer such knowledge~\citep{hinton2015distilling,romero2015fitnets,gou2021knowledge}. However, conventional distillation is not well suited to diffusion-to-DFSR transfer because the teacher and student follow very different generation processes. Feature-level distillation usually relies on meaningful correspondence between intermediate representations~\citep{romero2015fitnets,he2019knowledge}, whereas diffusion features depend on noisy states and denoising dynamics, while DFSR features are optimized for deterministic reconstruction.
\begin{wrapfigure}{r}{0.48\textwidth}
\vspace{-6pt}
\centering
\includegraphics[width=\linewidth]{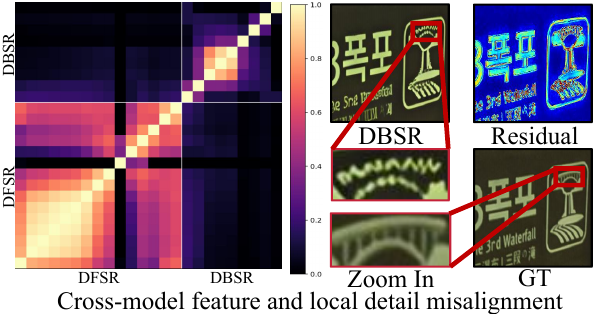}
\caption{
\textbf{Motivation.}
CKA~\citep{kornblith2019similarity} (left) reveals weak feature correspondence between DBSR and DFSR, while residual and zoom-in comparisons (right) show local discrepancies between DBSR outputs and paired HR targets.
}
\label{fig:motivation}
\vspace{-12pt}
\end{wrapfigure}
As shown in Fig.~\ref{fig:motivation}, DBSR and DFSR exhibit limited feature correspondence. Output distillation also has an inherent limitation: diffusion-based restoration may generate perceptually plausible details that do not exactly agree with the paired HR target, and directly imitating them may transfer stochastic or teacher-specific textures. Recent work such as DTKD~\citep{park2026dtkd} transfers diffusion knowledge to feed-forward students through output supervision. In contrast, we explore whether the diffusion model can act as a \emph{distribution-level teacher}, avoiding both feature alignment and imitation of sampled SR outputs.

\begin{figure}[t]
\centering
\includegraphics[width=\linewidth]{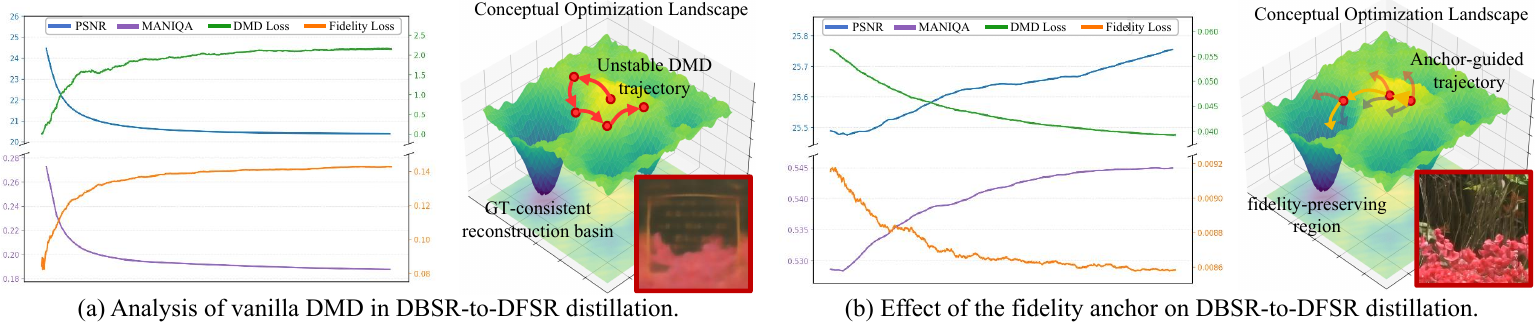}
\caption{
\textbf{Effect of fidelity anchoring on distribution matching.}
(a) Vanilla DMD causes unstable optimization and rapid PSNR degradation when directly applied to SR.
(b) Sample-level fidelity supervision stabilizes distribution matching and preserves reconstruction quality.
The optimization landscapes are conceptual illustrations.
}
\label{fig:vanilla_dmd_failure}
\vspace{-6pt}
\end{figure}

Distribution Matching Distillation (DMD)~\citep{yin2024one,yin2024improved} is particularly suitable for this purpose because real--fake score differences can supervise a differentiable generator without requiring architectural correspondence. This makes it possible to use a pretrained diffusion model to train independently designed DFSR networks such as SwinIR, HAT, and SeeMoRe. However, directly applying DMD to SR is insufficient. SR is a paired restoration task in which each LR input should correspond to a specific HR target, whereas distribution matching mainly constrains the overall output distribution. As shown in Fig.~\ref{fig:vanilla_dmd_failure}(a), vanilla DMD causes unstable optimization and rapid PSNR degradation. Paired reconstruction supervision substantially stabilizes training and prevents the diffusion prior from overriding input-specific content (Fig.~\ref{fig:vanilla_dmd_failure}(b)). Thus, distribution-level diffusion distillation requires explicit adaptation to the fidelity-sensitive nature of SR.

To this end, we propose \textbf{PhoenixSR}, a generative heterogeneous distillation framework that transfers a pretrained diffusion prior to independently designed diffusion-free SR networks through fidelity-aware distribution matching. During training, DFSR outputs are mapped to the diffusion latent space, where real and fake score models provide distribution-level guidance, while paired SR supervision preserves LR--HR correspondence. We introduce \textbf{Heterogeneous Distribution Adaptation (HDA)} to adapt score estimation to the SR training distribution and stabilize the evolving student distribution. We further employ \textbf{Directional Reliability Weighting (DRW)}, a lightweight sample reweighting strategy based on denoising-residual direction consistency, to reduce unstable score guidance. All diffusion-related components are removed after distillation, leaving the original DFSR architecture and inference cost unchanged.

The contributions are summarized as follows:
\vspace{-7pt}
\begin{itemize}
\setlength{\itemsep}{1pt}
\setlength{\parskip}{0pt}
\setlength{\parsep}{0pt}
\setlength{\topsep}{2pt}

\item To the best of our knowledge, PhoenixSR is the first framework that applies distribution matching distillation to transfer a pretrained diffusion generative prior into independently designed diffusion-free SR networks, without modifying their inference architecture.

\item We develop a fidelity-aware distribution distillation scheme for paired SR. HDA adapts real- and fake-score estimation to the SR training distribution and combines distribution-level diffusion guidance with paired reconstruction supervision, while DRW further improves training stability through lightweight sample reweighting.

\item Extensive experiments on three SR benchmarks and six CNN, Transformer, and lightweight SR backbones show that PhoenixSR consistently improves perceptual quality while largely preserving reconstruction fidelity and the original inference cost. 
\vspace{-6pt}

\end{itemize}

\section{Related Work}

We summarize the most relevant work here and defer detailed discussion to Appendix~\ref{app:related_work}. Real-world SR has evolved from CNN/Transformer reconstruction models~\citep{dong2014learning,kim2016accurate,lim2017enhanced,zhang2018rcan,liang2021swinir,chen2023hat} to diffusion-based methods that exploit strong generative priors~\citep{wang2023stablesr,yue2024resshift,lin2024diffbir,wu2024seesr,yu2024scaling}, while efficient SR focuses on lightweight feed-forward inference~\citep{kong2022rlfn,sun2023safmn,zamfir2024seemore}. Conventional knowledge distillation transfers outputs or intermediate representations~\citep{hinton2015distilling,romero2015fitnets,he2019knowledge,gou2021knowledge}, whereas most diffusion distillation methods accelerate sampling while retaining diffusion-derived generators~\citep{salimans2022progressive,song2023consistency,meng2023distillation}. DMD/DMD2 instead perform distribution-level distillation through diffusion-score differences~\citep{yin2024one,yin2024improved}, and related SDS/VSD methods also use pretrained diffusion scores as supervision~\citep{poole2023dreamfusion,wang2023prolificdreamer}. Recent work has explored heterogeneous diffusion transfer in restoration: DTKD distills a diffusion SR teacher into a Transformer student through frequency-aware output supervision~\citep{park2026dtkd}, DM-SR applies distribution matching to SR while retaining a diffusion denoiser at inference~\citep{park2026dmsr}, and DUO-VSR applies DMD to a diffusion-derived one-step VSR generator~\citep{lv2026duovsr}. Cross-space distillation further studies distribution transfer across different diffusion latent spaces~\citep{nguyen2026crossspace}. In contrast, PhoenixSR transfers a pretrained diffusion prior through distribution matching into independently designed diffusion-free SR networks, using paired supervision to preserve reconstruction fidelity and removing all diffusion components after training.

\section{Method}
\label{sec:method}

\paragraph{Problem Formulation.}
Let $\mathbf{y}$ and $\mathbf{x}$ denote a degraded low-resolution (LR) image and its paired high-resolution (HR) target, respectively. A diffusion-free SR (DFSR) student $G_\theta$ predicts $\hat{\mathbf{x}}_0=G_\theta(\mathbf{y})$. Since the diffusion teacher operates in Stable Diffusion latent space, its frozen VAE encoder $E$ maps the student output to
\begin{equation}
\hat{\mathbf{x}}_0=G_\theta(\mathbf{y}),\qquad
\hat{\mathbf{z}}_0=E(\hat{\mathbf{x}}_0),\qquad
\hat{\mathbf{z}}_t=\alpha_t\hat{\mathbf{z}}_0+\sigma_t\boldsymbol{\epsilon},
\quad \boldsymbol{\epsilon}\sim\mathcal{N}(\mathbf{0},\mathbf{I}),
\label{eq:problem}
\end{equation}
where $\hat{\mathbf{z}}_t$ is the noisy student latent at timestep $t$. Our goal is to transfer the generative prior of a pretrained diffusion teacher to an independently designed DFSR student while preserving its original feed-forward inference architecture.

Following Distribution Matching Distillation (DMD)~\citep{yin2024one,yin2024improved}, a real-score model characterizes the target distribution, while an online fake-score model tracks the evolving student distribution. In score notation, the distribution-matching gradient is
\begin{equation}
\nabla_\theta\mathcal{L}_{\mathrm{DMD}}
=
\mathbb{E}\!\left[
w(t)\big(
\mathbf{s}_{\mathrm{fake}}(\hat{\mathbf{z}}_t,t,\mathbf{c})-
\mathbf{s}_{\mathrm{real}}(\hat{\mathbf{z}}_t,t,\mathbf{c})
\big)
\frac{\partial\hat{\mathbf{z}}_0}{\partial\theta}
\right],
\label{eq:dmd_main}
\end{equation}
where $\mathbf{c}$ denotes the diffusion text condition and $w(t)$ absorbs timestep-dependent factors. The reported setting uses a fixed quality prompt, and the implementation realizes the score difference through the equivalent denoised-latent parameterization with per-sample normalization for stability. Importantly, DMD provides distribution-level supervision without requiring architectural correspondence between the diffusion model and the student, making it suitable for transferring diffusion priors to independently designed DFSR networks. However, distribution matching alone does not enforce consistency with the paired HR target. Preliminaries are provided in the supplementary material.

\subsection{Motivation}
\label{sec:motivation_method}

The architecture-independent nature of DMD provides a natural way to transfer diffusion generative priors to native DFSR networks. However, applying it to paired SR introduces two practical challenges. First, the real-score model should provide suitable guidance for the SR training domain, while the online fake-score model must continuously track the evolving student distribution. Second, unlike unconditional generation, SR requires each output to remain faithful to its corresponding LR--HR pair. As shown in Fig.~\ref{fig:vanilla_dmd_failure}, vanilla DMD causes unstable optimization and substantial PSNR degradation, whereas paired reconstruction supervision stabilizes the transfer. We therefore introduce \textbf{Heterogeneous Distribution Adaptation (HDA)} for score adaptation and fidelity anchoring, together with \textbf{Directional Reliability Weighting (DRW)} as an auxiliary sample reweighting strategy.

\subsection{Heterogeneous Distribution Adaptation}
\label{sec:hda}

\begin{figure}[t]
    \centering
    \includegraphics[width=\linewidth]{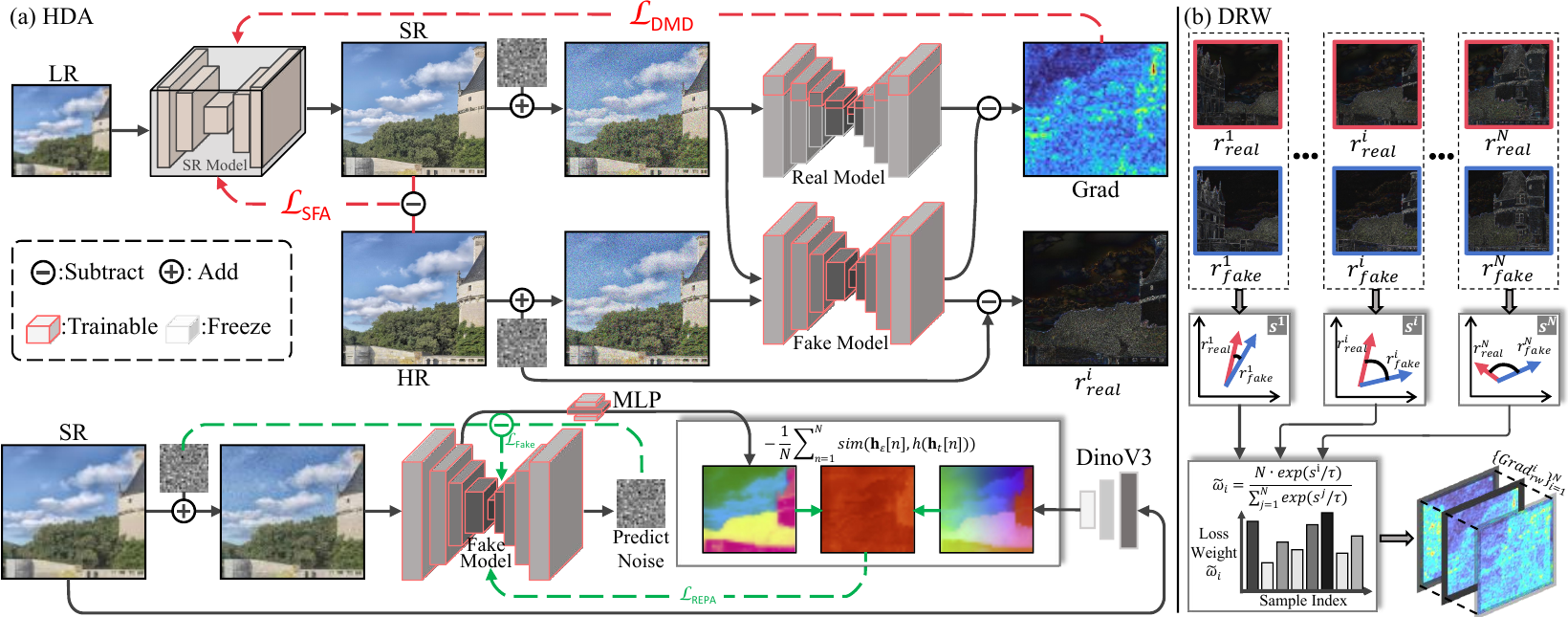}
    \caption{
    \textbf{Overview of PhoenixSR.}
    HDA adapts real-score estimation, tracks the evolving student distribution, and anchors paired fidelity. DRW reweights sample-wise guidance according to residual-direction consistency.
    }
    \label{fig:hda_framework}
    \vspace{-10pt}
\end{figure}

Standard DMD uses a fixed real-score model for the target distribution and an online fake-score model for the generator distribution. Under diffusion-to-DFSR transfer, the real-score model needs to better match the SR training domain, while the fake-score model must track a continuously changing student distribution. Moreover, distribution matching alone does not guarantee paired reconstruction fidelity. HDA addresses these issues with an adapted real-score model, a representation-regularized fake-score model, and a sample-level fidelity anchor.

\paragraph{LoRA-adapted real-score model.}
The pretrained diffusion teacher learns a general natural-image prior, while the HR images used for real-world SR exhibit task-specific image statistics. Directly using the frozen score model may therefore provide suboptimal guidance for SR distillation.

To improve target-score compatibility while preserving the pretrained prior, we insert lightweight LoRA adapters~\citep{hu2022lora} into the real-score model and freeze its original backbone. Given an HR target $\mathbf{x}$, let $\mathbf{z}_0^{HR}=E(\mathbf{x})$ and
\begin{equation}
\mathbf{z}_t^{HR}=\alpha_t\mathbf{z}_0^{HR}+\sigma_t\boldsymbol{\epsilon},
\qquad
\mathcal{L}_{\mathrm{real}}
=
\mathbb{E}\!\left[
\left\|
\boldsymbol{\epsilon}_{\mathrm{real}}(\mathbf{z}_t^{HR},t,\mathbf{c})-\boldsymbol{\epsilon}
\right\|_2^2
\right].
\label{eq:hda_real_main}
\end{equation}
Only the low-rank adapters are optimized, limiting adaptation while retaining the pretrained backbone. Detailed LoRA formulation and analysis are provided in the supplementary material.

\paragraph{Representation-regularized fake-score model.}
The fake-score model estimates the evolving student distribution and is fully optimized online using noisy student outputs. Since the student output distribution changes continuously during distillation, directly fitting the fake-score model may make it sensitive to transient artifacts. We therefore regularize its intermediate representations following REPA~\citep{yu2025representation}, aligning fake-UNet features with a frozen DINOv3 encoder~\citep{simeoni2025dinov3}:
\begin{equation}
\mathcal{L}_{\mathrm{fake}}
=
\mathbb{E}\!\left[
\left\|
\boldsymbol{\epsilon}_{\mathrm{fake}}(\hat{\mathbf{z}}_t,t,\mathbf{c})-\boldsymbol{\epsilon}
\right\|_2^2
\right],
\qquad
\mathcal{L}_{\mathrm{fake}}^{\mathrm{total}}
=
\mathcal{L}_{\mathrm{fake}}
+
\lambda_{\mathrm{repa}}\mathcal{L}_{\mathrm{REPA}}.
\label{eq:hda_fake_main}
\end{equation}
This regularizes the fake-score representation while tracking the evolving student distribution. The full REPA objective is deferred to the supplementary material.

\paragraph{Sample-level fidelity anchor.}
Reliable score estimation alone is insufficient because DMD remains a distribution-level objective. For paired SR, the student should approach a high-quality image distribution while remaining faithful to the specific LR--HR pair. We therefore introduce a sample-level fidelity anchor (SFA):
\begin{equation}
\mathcal{L}_{\mathrm{SFA}}
=
\lambda_{\mathrm{pix}}\mathcal{L}_{\mathrm{pix}}
+
\lambda_{\mathrm{per}}\mathcal{L}_{\mathrm{per}}
+
\lambda_{\mathrm{adv}}\mathcal{L}_{\mathrm{adv}},
\qquad
\nabla_\theta\mathcal{L}_{G}
=
\lambda_{\mathrm{dmd}}\nabla_\theta\mathcal{L}_{\mathrm{DMD}}
+
\nabla_\theta\mathcal{L}_{\mathrm{SFA}}.
\label{eq:hda_main}
\end{equation}

\begin{wrapfigure}{r}{0.48\textwidth}
    \vspace{-8pt}
    \centering
    \includegraphics[width=\linewidth]{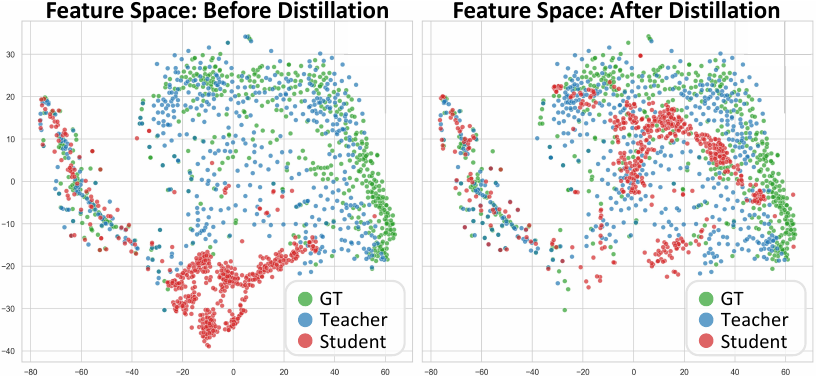}
    \caption{
    \textbf{t-SNE projection of feature representations.}
    After distillation, DFSR representations move toward the region between real HR samples and DBSR outputs in t-SNE projection.
    }
    \label{fig:tsne_hda}
    \vspace{-35pt}
\end{wrapfigure}

The pixel and perceptual terms preserve reconstruction fidelity, while the adversarial term promotes realistic textures. As illustrated in Fig.~\ref{fig:vanilla_dmd_failure}(b), SFA constrains distribution matching with paired reconstruction supervision and substantially stabilizes training.

Together, LoRA adaptation improves target-score compatibility, representation-regularized online optimization improves student-distribution tracking, and SFA preserves input-specific reconstruction. These components enable distribution-level diffusion distillation to operate effectively on independently designed DFSR students.

As a qualitative representation-space diagnostic, Fig.~\ref{fig:tsne_hda} visualizes DFSR outputs before and after distillation together with real HR samples and DBSR outputs using t-SNE~\citep{vandermaaten2008tsne}. After distillation, the projected DFSR representations move toward the intermediate region between real HR and DBSR outputs rather than directly overlapping the DBSR cluster, consistent with transferring diffusion priors while retaining HR-oriented reconstruction characteristics.

\subsection{Directional Reliability Weighting}
\label{sec:drw}

Although HDA improves real- and fake-score estimation, the online fake-score model may behave differently across samples as the student distribution continuously evolves. Samples that are not yet well represented by the fake-score model can therefore produce less stable distributional guidance. We use \textbf{Directional Reliability Weighting (DRW)} to adjust their relative contributions during training.

For each pair, the generated output and HR target are encoded as $\mathbf{z}_{0}^{G,i}$ and $\mathbf{z}_{0}^{HR,i}$ and perturbed using the same timestep $t$ and Gaussian noise $\boldsymbol{\epsilon}^{i}$:
\begin{equation}
\mathbf{z}_{t}^{G,i}=\alpha_t\mathbf{z}_{0}^{G,i}+\sigma_t\boldsymbol{\epsilon}^{i},
\qquad
\mathbf{z}_{t}^{HR,i}=\alpha_t\mathbf{z}_{0}^{HR,i}+\sigma_t\boldsymbol{\epsilon}^{i}.
\label{eq:drw_noise}
\end{equation}
Using the same perturbation removes variation caused purely by timestep or sampled noise, allowing the fake-score responses around the generated and HR latents to be compared under the same condition.

We evaluate both noisy states using the fake-score model and compute their denoising residuals and directional consistency:
\begin{equation}
\mathbf{r}_{G}^{i}
=
\boldsymbol{\epsilon}_{\mathrm{fake}}(\mathbf{z}_{t}^{G,i},t,\mathbf{c}^{i})
-\boldsymbol{\epsilon}^{i},
\quad
\mathbf{r}_{HR}^{i}
=
\boldsymbol{\epsilon}_{\mathrm{fake}}(\mathbf{z}_{t}^{HR,i},t,\mathbf{c}^{i})
-\boldsymbol{\epsilon}^{i},
\quad
s_i=\cos(\mathbf{r}_{G}^{i},\mathbf{r}_{HR}^{i}).
\label{eq:drw_score}
\end{equation}
Here $\mathbf{c}^{i}$ denotes the diffusion condition. We use the residual-direction consistency as a sample-wise proxy for the local behavior of the fake-score model rather than as an image-quality measure. Higher $s_i$ indicates more similar denoising behavior around the generated and HR regions.

We convert $s_i$ into a normalized reliability weight using a temperature-controlled softmax and clip extreme values for stability:
\begin{equation}
\omega_i
=
\mathrm{clip}\!\left(
N\frac{\exp(s_i/\tau)}{\sum_{j=1}^{N}\exp(s_j/\tau)},
\omega_{\min},\omega_{\max}
\right),
\qquad
\mathcal{L}_{\mathrm{DMD}}^{\mathrm{DRW}}
=
\frac{1}{N}\sum_{i=1}^{N}
\operatorname{sg}(\omega_i)\mathcal{L}_{\mathrm{DMD}}^{i}.
\label{eq:drw_main}
\end{equation}
Here $N$ is the number of samples in the reliability-normalization pool; in distributed training, similarities are gathered across workers before the softmax. The factor $N$ places the softmax weights on a unit-mean scale before clipping, while $\tau$ controls the contrast among samples. Here $\operatorname{sg}(\cdot)$ denotes stop-gradient; no gradient is propagated through $s_i$ or $\omega_i$ during the generator update. The resulting weights are used to adjust the relative contribution of different samples to the DMD objective.

The timestep schedule, teacher warm-up, asynchronous optimization, and complete PhoenixSR training procedure are provided in the supplementary material.

\section{Experiments}
\label{sec:Experiments}
\paragraph{Implementation.}
Following the real-world SR setting of SwinIR~\citep{liang2021swinir}, we construct the HR training set from DIV2K~\citep{agustsson2017ntire}, Flickr2K~\citep{lim2017enhanced}, OST~\citep{wang2018recovering}, WED~\citep{ma2017waterloo}, FFHQ~\citep{karras2019style}, Manga109~\citep{matsui2017sketch}, and SCUT-CTW1500~\citep{liu2019curved}, with LR inputs synthesized using the Real-ESRGAN degradation pipeline~\citep{wang2021realesrgan}. We evaluate $4\times$ SR on RealSR~\citep{cai2019toward}, DRealSR~\citep{wei2020component}, and DIV2K-Val~\citep{agustsson2017ntire}, using PSNR~\citep{hore2010image}, LPIPS~\citep{zhang2018unreasonable}, MANIQA~\citep{yang2022maniqa}, NIQE~\citep{mittal2013making}, MUSIQ~\citep{ke2021musiq}, CLIP-IQA~\citep{wang2023exploring}, LIQE-Mix~\citep{zhang2023blind}, ARNIQA~\citep{agnolucci2024arniqa}, and TOPIQ~\citep{chen2024topiq}. PhoenixSR is applied to SwinIR-L/M/S~\citep{liang2021swinir}, HAT~\citep{chen2023hat}, Real-ESRGAN~\citep{wang2021realesrgan}, and SeeMoRe~\citep{zamfir2024seemore}, using pretrained Stable Diffusion~2.1~\citep{rombach2022high} as the training-only diffusion teacher with a fixed quality prompt; the real-score model is LoRA-adapted~\citep{hu2022lora}, while the fake-score model is optimized online with REPA regularization~\citep{yu2025representation}. Quantitative comparisons include StableSR~\citep{wang2023stablesr}, SeeSR~\citep{wu2024seesr}, SinSR~\citep{wang2024sinsr}, OSEDiff~\citep{wu2024osediff}, and AdcSR~\citep{chen2025adcsr}, with additional diffusion-based methods shown qualitatively. All auxiliary diffusion components are discarded after distillation, leaving the student inference architecture unchanged; detailed optimization and hyperparameter settings are provided in the supplementary material.


\begin{table*}[t]
\centering
\vspace{-2mm}
\caption{
Quantitative comparison on RealSR. PhoenixSR is applied to DFSR backbones, while DBSR methods are included as reference baselines. Gray cells denote results with PhoenixSR; red and blue denote performance improvement and degradation, respectively.
}
\label{tab:main_realsr}

\scriptsize
\setlength{\tabcolsep}{3.2pt}
\renewcommand{\arraystretch}{0.80}
\setlength{\aboverulesep}{0.2ex}
\setlength{\belowrulesep}{0.2ex}

\resizebox{\linewidth}{!}{
\begin{tabular}{
c|
ccccccccc|
@{\hspace{2.0pt}}c
@{\hspace{2.5pt}}c
@{\hspace{2.5pt}}c
}
\toprule

\multirow{2}{*}{Model}
& \multirow{2}{*}{PSNR$\uparrow$}
& \multirow{2}{*}{LPIPS$\downarrow$}
& \multirow{2}{*}{MANIQA$\uparrow$}
& \multirow{2}{*}{NIQE$\downarrow$}
& \multirow{2}{*}{MUSIQ$\uparrow$}
& \multirow{2}{*}{CLIP-IQA$\uparrow$}
& \multirow{2}{*}{LIQE-Mix$\uparrow$}
& \multirow{2}{*}{ARNIQA$\uparrow$}
& \multirow{2}{*}{TOPIQ$\uparrow$}
& Params
& FLOPs
& Runtime \\

& & & & & & & & & &
{\tiny (M)}
& {\tiny (G)}
& {\tiny (ms)} \\

\midrule

\multicolumn{13}{c}{\textit{DBSR Models}} \\
\midrule

StableSR
& 24.4272
& 0.2636
& 0.5910
& 6.5779
& 61.8041
& 0.5644
& 3.5794
& 0.6772
& 0.5121
& 1410
& 162731
& 17870 \\

SeeSR
& 25.1475
& 0.3007
& 0.6451
& 5.3958
& 69.8181
& 0.6703
& 3.6576
& 0.7152
& 0.6888
& 2511
& 131980
& 5451 \\

SinSR
& 26.1309
& 0.3068
& 0.5432
& 6.1038
& 61.2905
& 0.6236
& 3.5595
& 0.6601
& 0.5309
& 119
& 5392
& 201 \\

OSEDiff
& 25.1554
& 0.2928
& 0.6343
& 5.6243
& 69.1793
& 0.6669
& 4.1382
& 0.6935
& 0.6292
& 1761
& 4611
& 169 \\

AdcSR
& 24.4016
& 0.2939
& 0.6372
& 5.2979
& 70.0261
& 0.6939
& 4.2198
& 0.7227
& 0.6863
& 456
& 990
& 101 \\

\midrule
\multicolumn{13}{c}{\textit{DFSR Models}} \\
\midrule


SwinIR-L
& 26.3071
& 0.2531
& 0.5228
& 5.7025
& 58.6388
& 0.4502
& 3.0465
& 0.6593
& 0.4774
& \multirow{3}{*}{28.01}
& \multirow{3}{*}{1006.56}
& \multirow{3}{*}{202} \\

\cellcolor{phoenixgray}{\scriptsize (w/PhoenixSR)}
& \cellcolor{phoenixgray}26.7437
& \cellcolor{phoenixgray}0.2500
& \cellcolor{phoenixgray}0.5508
& \cellcolor{phoenixgray}5.4432
& \cellcolor{phoenixgray}61.3691
& \cellcolor{phoenixgray}0.4824
& \cellcolor{phoenixgray}3.1735
& \cellcolor{phoenixgray}0.6638
& \cellcolor{phoenixgray}0.5063
& & & \\

{\tiny Gain}
& \posgain{+0.4366}
& \posgain{-0.0031}
& \posgain{+0.0280}
& \posgain{-0.2593}
& \posgain{+2.7303}
& \posgain{+0.0322}
& \posgain{+0.1270}
& \posgain{+0.0045}
& \posgain{+0.0289}
& & & \\

\cmidrule{1-13}


HAT
& 25.9792
& 0.2421
& 0.5517
& 6.1520
& 57.9440
& 0.4068
& 3.1663
& 0.6552
& 0.4660
& \multirow{3}{*}{20.77}
& \multirow{3}{*}{833.45}
& \multirow{3}{*}{201} \\

\cellcolor{phoenixgray}{\scriptsize (w/PhoenixSR)}
& \cellcolor{phoenixgray}26.1969
& \cellcolor{phoenixgray}0.2399
& \cellcolor{phoenixgray}0.5708
& \cellcolor{phoenixgray}5.9102
& \cellcolor{phoenixgray}61.1100
& \cellcolor{phoenixgray}0.4599
& \cellcolor{phoenixgray}3.4304
& \cellcolor{phoenixgray}0.6732
& \cellcolor{phoenixgray}0.5073
& & & \\

{\tiny Gain}
& \posgain{+0.2177}
& \posgain{-0.0022}
& \posgain{+0.0191}
& \posgain{-0.2418}
& \posgain{+3.1660}
& \posgain{+0.0531}
& \posgain{+0.2641}
& \posgain{+0.0180}
& \posgain{+0.0413}
& & & \\

\cmidrule{1-13}


Real-ESRGAN
& 25.6900
& 0.2708
& 0.5505
& 5.7937
& 60.3978
& 0.4761
& 3.3976
& 0.6754
& 0.5154
& \multirow{3}{*}{16.70}
& \multirow{3}{*}{587.62}
& \multirow{3}{*}{50} \\

\cellcolor{phoenixgray}{\scriptsize (w/PhoenixSR)}
& \cellcolor{phoenixgray}25.9452
& \cellcolor{phoenixgray}0.2626
& \cellcolor{phoenixgray}0.5590
& \cellcolor{phoenixgray}5.6776
& \cellcolor{phoenixgray}61.6747
& \cellcolor{phoenixgray}0.4740
& \cellcolor{phoenixgray}3.6010
& \cellcolor{phoenixgray}0.6827
& \cellcolor{phoenixgray}0.5302
& & & \\

{\tiny Gain}
& \posgain{+0.2552}
& \posgain{-0.0082}
& \posgain{+0.0085}
& \posgain{-0.1161}
& \posgain{+1.2769}
& \neggain{-0.0021}
& \posgain{+0.2034}
& \posgain{+0.0073}
& \posgain{+0.0148}
& & & \\

\cmidrule{1-13}


SwinIR-M
& 25.8912
& 0.2608
& 0.5226
& 5.7635
& 59.1610
& 0.4435
& 3.1377
& 0.6634
& 0.4860
& \multirow{3}{*}{11.72}
& \multirow{3}{*}{448.57}
& \multirow{3}{*}{111} \\

\cellcolor{phoenixgray}{\scriptsize (w/PhoenixSR)}
& \cellcolor{phoenixgray}26.3384
& \cellcolor{phoenixgray}0.2593
& \cellcolor{phoenixgray}0.5572
& \cellcolor{phoenixgray}5.4158
& \cellcolor{phoenixgray}62.1048
& \cellcolor{phoenixgray}0.5112
& \cellcolor{phoenixgray}3.2828
& \cellcolor{phoenixgray}0.6730
& \cellcolor{phoenixgray}0.5182
& & & \\

{\tiny Gain}
& \posgain{+0.4472}
& \posgain{-0.0015}
& \posgain{+0.0346}
& \posgain{-0.3477}
& \posgain{+2.9438}
& \posgain{+0.0677}
& \posgain{+0.1451}
& \posgain{+0.0096}
& \posgain{+0.0322}
& & & \\

\cmidrule{1-13}


SwinIR-S
& 25.3030
& 0.2965
& 0.4413
& 7.4007
& 54.2403
& 0.4463
& 2.7434
& 0.6502
& 0.4458
& \multirow{3}{*}{1.05}
& \multirow{3}{*}{79.95}
& \multirow{3}{*}{61} \\

\cellcolor{phoenixgray}{\scriptsize (w/PhoenixSR)}
& \cellcolor{phoenixgray}25.2573
& \cellcolor{phoenixgray}0.3035
& \cellcolor{phoenixgray}0.5529
& \cellcolor{phoenixgray}5.5930
& \cellcolor{phoenixgray}65.3453
& \cellcolor{phoenixgray}0.5790
& \cellcolor{phoenixgray}3.9146
& \cellcolor{phoenixgray}0.6957
& \cellcolor{phoenixgray}0.5851
& & & \\

{\tiny Gain}
& \neggain{-0.0457}
& \neggain{+0.0070}
& \posgain{+0.1116}
& \posgain{-1.8077}
& \posgain{+11.1050}
& \posgain{+0.1327}
& \posgain{+1.1712}
& \posgain{+0.0455}
& \posgain{+0.1393}
& & & \\

\cmidrule{1-13}


SeeMoRe
& 26.0739
& 0.2953
& 0.3476
& 8.6255
& 48.7427
& 0.3813
& 1.9931
& 0.4839
& 0.3151
& \multirow{3}{*}{0.97}
& \multirow{3}{*}{28.55}
& \multirow{3}{*}{40} \\

\cellcolor{phoenixgray}{\scriptsize (w/PhoenixSR)}
& \cellcolor{phoenixgray}25.9329
& \cellcolor{phoenixgray}0.2663
& \cellcolor{phoenixgray}0.5454
& \cellcolor{phoenixgray}5.1620
& \cellcolor{phoenixgray}62.4174
& \cellcolor{phoenixgray}0.5283
& \cellcolor{phoenixgray}3.6825
& \cellcolor{phoenixgray}0.6757
& \cellcolor{phoenixgray}0.4976
& & & \\

{\tiny Gain}
& \neggain{-0.1410}
& \posgain{-0.0290}
& \posgain{+0.1978}
& \posgain{-3.4635}
& \posgain{+13.6747}
& \posgain{+0.1470}
& \posgain{+1.6894}
& \posgain{+0.1918}
& \posgain{+0.1825}
& & & \\

\bottomrule
\end{tabular}
}

\vspace{-2mm}
\end{table*}


\begin{table*}[t]
\centering
\vspace{-2mm}
\caption{
Quantitative comparison on DRealSR.
}
\label{tab:main_drealsr}

\scriptsize
\setlength{\tabcolsep}{5.0pt}
\renewcommand{\arraystretch}{0.80}
\setlength{\aboverulesep}{0.2ex}
\setlength{\belowrulesep}{0.2ex}

\begin{tabular}{c|ccccccccc}
\toprule
Model
& PSNR$\uparrow$
& LPIPS$\downarrow$
& MANIQA$\uparrow$
& NIQE$\downarrow$
& MUSIQ$\uparrow$
& CLIP-IQA$\uparrow$
& LIQE-Mix$\uparrow$
& ARNIQA$\uparrow$
& TOPIQ$\uparrow$ \\
\midrule

\multicolumn{10}{c}{\textit{DBSR Models}} \\
\midrule

StableSR
& 28.4070 & 0.2726 & 0.4914 & 7.5141 & 51.4058
& 0.5046 & 2.6971 & 0.5762 & 0.4321 \\

SeeSR
& 28.0712 & 0.3174 & 0.6052 & 6.4061 & 65.0873
& 0.6909 & 3.6846 & 0.6880 & 0.6572 \\

SinSR
& 28.2012 & 0.3511 & 0.5034 & 6.7220 & 57.5658
& 0.6574 & 3.5766 & 0.6224 & 0.5288 \\

OSEDiff
& 27.9321 & 0.2971 & 0.5874 & 6.5227 & 64.8184
& 0.6909 & 3.9768 & 0.6702 & 0.6005 \\

AdcSR
& 26.2041 & 0.3156 & 0.5937 & 6.1738 & 66.8444
& 0.7206 & 4.2538 & 0.6985 & 0.6602 \\

\midrule
\multicolumn{10}{c}{\textit{DFSR Models}} \\
\midrule

SwinIR-L
& 28.4999 & 0.2739 & 0.4751 & 6.6093 & 52.8383
& 0.4680 & 2.9287 & 0.5935 & 0.4425 \\

\rowcolor{phoenixgray}
{\scriptsize (w/PhoenixSR)}
& 29.1477 & 0.2721 & 0.5095 & 6.1132 & 54.6482
& 0.5006 & 3.1093 & 0.6045 & 0.4663 \\

{\tiny Gain}
& \posgain{+0.6478}
& \posgain{-0.0018}
& \posgain{+0.0344}
& \posgain{-0.4961}
& \posgain{+1.8099}
& \posgain{+0.0326}
& \posgain{+0.1806}
& \posgain{+0.0110}
& \posgain{+0.0238} \\

\cmidrule{1-10}

HAT
& 28.9684 & 0.2594 & 0.4887 & 7.1056 & 50.8062
& 0.4196 & 2.8382 & 0.5710 & 0.4183 \\

\rowcolor{phoenixgray}
{\scriptsize (w/PhoenixSR)}
& 29.0597 & 0.2546 & 0.5134 & 6.4380 & 54.1864
& 0.4735 & 3.1686 & 0.6105 & 0.4510 \\

{\tiny Gain}
& \posgain{+0.0913}
& \posgain{-0.0048}
& \posgain{+0.0247}
& \posgain{-0.6676}
& \posgain{+3.3802}
& \posgain{+0.0539}
& \posgain{+0.3304}
& \posgain{+0.0395}
& \posgain{+0.0327} \\

\cmidrule{1-10}

Real-ESRGAN
& 28.6276 & 0.2819 & 0.4903 & 6.7022 & 54.2878
& 0.4812 & 3.1287 & 0.6103 & 0.4627 \\

\rowcolor{phoenixgray}
{\scriptsize (w/PhoenixSR)}
& 28.7265 & 0.2790 & 0.5041 & 6.3026 & 55.8058
& 0.5072 & 3.3049 & 0.6282 & 0.4771 \\

{\tiny Gain}
& \posgain{+0.0989}
& \posgain{-0.0029}
& \posgain{+0.0138}
& \posgain{-0.3996}
& \posgain{+1.5180}
& \posgain{+0.0260}
& \posgain{+0.1762}
& \posgain{+0.0179}
& \posgain{+0.0144} \\

\cmidrule{1-10}

SwinIR-M
& 28.1926 & 0.2827 & 0.4672 & 6.5990 & 53.1293
& 0.4517 & 2.9558 & 0.6054 & 0.4518 \\

\rowcolor{phoenixgray}
{\scriptsize (w/PhoenixSR)}
& 28.7346 & 0.2811 & 0.5076 & 5.9858 & 55.3788
& 0.5172 & 3.1711 & 0.6248 & 0.4793 \\

{\tiny Gain}
& \posgain{+0.5420}
& \posgain{-0.0016}
& \posgain{+0.0404}
& \posgain{-0.6132}
& \posgain{+2.2495}
& \posgain{+0.0655}
& \posgain{+0.2153}
& \posgain{+0.0194}
& \posgain{+0.0275} \\

\cmidrule{1-10}

SwinIR-S
& 28.5549 & 0.3044 & 0.4065 & 8.5537 & 49.5158
& 0.4196 & 2.5339 & 0.5988 & 0.4223 \\

\rowcolor{phoenixgray}
{\scriptsize (w/PhoenixSR)}
& 28.3787 & 0.3125 & 0.4985 & 6.3289 & 59.6194
& 0.5719 & 3.6983 & 0.6641 & 0.5341 \\

{\tiny Gain}
& \neggain{-0.1762}
& \neggain{+0.0081}
& \posgain{+0.0920}
& \posgain{-2.2248}
& \posgain{+10.1036}
& \posgain{+0.1523}
& \posgain{+1.1644}
& \posgain{+0.0653}
& \posgain{+0.1118} \\

\cmidrule{1-10}

SeeMoRe
& 28.9936 & 0.3264 & 0.2894 & 10.3603 & 39.9581
& 0.4225 & 1.6149 & 0.3754 & 0.2818 \\

\rowcolor{phoenixgray}
{\scriptsize (w/PhoenixSR)}
& 28.8592 & 0.2818 & 0.5060 & 6.1061 & 56.2313
& 0.5038 & 3.3956 & 0.6304 & 0.4696 \\

{\tiny Gain}
& \neggain{-0.1344}
& \posgain{-0.0446}
& \posgain{+0.2166}
& \posgain{-4.2542}
& \posgain{+16.2732}
& \posgain{+0.0813}
& \posgain{+1.7807}
& \posgain{+0.2550}
& \posgain{+0.1878} \\

\bottomrule
\end{tabular}

\vspace{-2mm}
\end{table*}

\subsection{Quantitative Comparison and Qualitative Results}
\label{sec:performance}

Tables~\ref{tab:main_realsr}, \ref{tab:main_drealsr}, and~\ref{tab:main_div2k_val} report quantitative results on RealSR, DRealSR, and DIV2K-Val, respectively. Across six CNN, Transformer, and lightweight DFSR backbones, PhoenixSR consistently improves perceptual quality while largely preserving reconstruction fidelity. For medium and large models, the improvement is generally well balanced. On RealSR, SwinIR-L, HAT, Real-ESRGAN, and SwinIR-M improve PSNR by $0.44$, $0.22$, $0.26$, and $0.45$ dB, respectively, together with improvements on most perceptual metrics. Similar trends are observed on DRealSR, where SwinIR-L and SwinIR-M gain $0.65$ and $0.54$ dB in PSNR while improving all reported no-reference metrics.

The gains are particularly noticeable for lightweight students. SwinIR-S ($1.05$M) and SeeMoRe ($0.97$M) remain within approximately $0.2$ dB of their original PSNR across the three datasets, while obtaining substantially better perceptual scores. For example, PhoenixSR improves MUSIQ by $11.11/10.10/17.73$ for SwinIR-S and $13.67/16.27/23.33$ for SeeMoRe on RealSR/DRealSR/DIV2K-Val, respectively. SeeMoRe further improves LPIPS on all three benchmarks. These results suggest that distribution-level diffusion guidance can provide useful generative priors even for highly compact feed-forward SR networks.

\begin{figure*}[t]
    \centering
    \includegraphics[width=\textwidth]{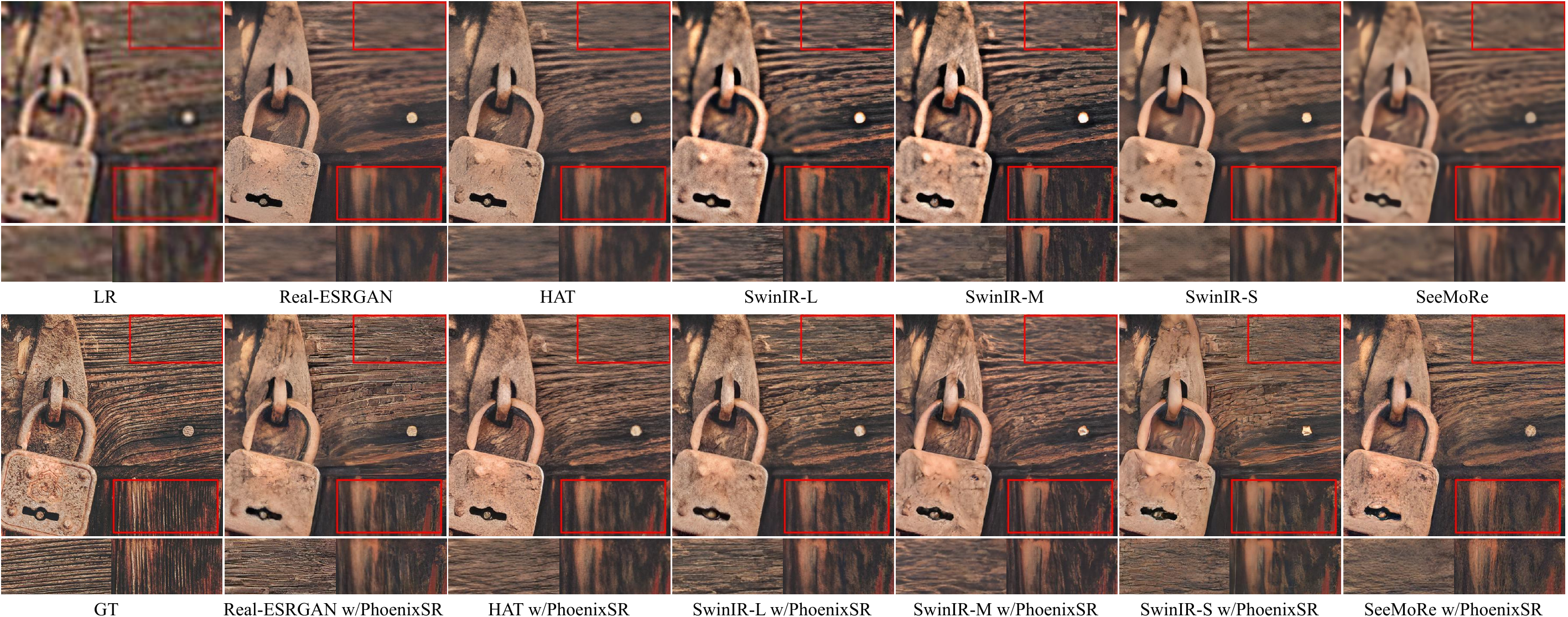}
    \caption{
    \textbf{Qualitative comparison on DIV2K-Val.}
    PhoenixSR enhances high-frequency details of diffusion-free SR models, producing clearer and more natural wood textures while preserving the underlying image structure.
    }
    \label{fig:visual_wood}
    \vspace{-15pt}
\end{figure*}

\begin{figure*}[t]
    \centering
    \includegraphics[width=\textwidth]{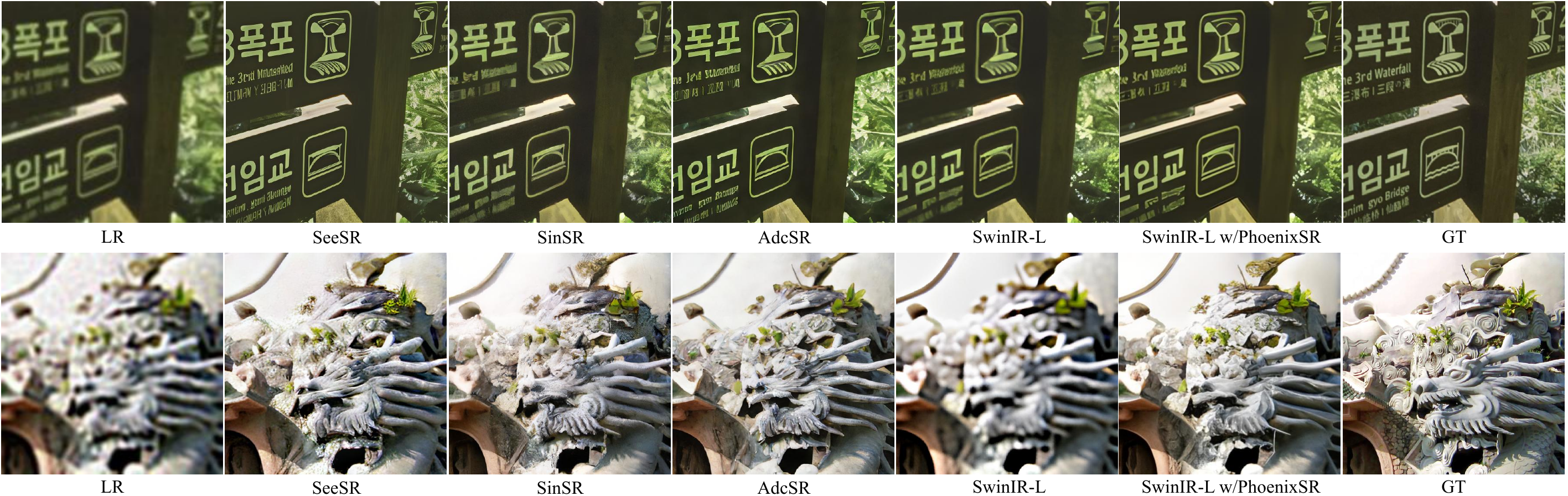}
    \caption{
    \textbf{Qualitative comparison with diffusion-based SR methods.}
Diffusion-based baselines may over-sharpen the statue surface or introduce hallucinated textures that deviate from the underlying geometry. PhoenixSR improves details while retaining more faithful structural reconstruction.
    }
    \label{fig:visual_diffusion_case}
\end{figure*}

Compared with DBSR baselines, PhoenixSR targets a different inference regime. StableSR, SeeSR, OSEDiff, and AdcSR retain diffusion-based or diffusion-derived generation pipelines at inference, whereas all diffusion-related components in PhoenixSR are removed after distillation. As shown in Table~\ref{tab:main_realsr}, the distilled students preserve the parameter count, FLOPs, and runtime of their original backbones. Figures~\ref{fig:visual_wood} and~\ref{fig:visual_diffusion_case} further show that PhoenixSR improves local textures while largely preserving the underlying image structure. The consistent behavior across different student architectures indicates that the proposed diffusion-prior transfer is not restricted to a particular DFSR backbone.


\begin{table*}[t]
\centering
\vspace{-2mm}
\caption{
Quantitative comparison on DIV2K-Val.
}
\label{tab:main_div2k_val}

\scriptsize
\setlength{\tabcolsep}{5.0pt}
\renewcommand{\arraystretch}{0.80}
\setlength{\aboverulesep}{0.2ex}
\setlength{\belowrulesep}{0.2ex}

\begin{tabular}{c|ccccccccc}
\toprule
Model
& PSNR$\uparrow$
& LPIPS$\downarrow$
& MANIQA$\uparrow$
& NIQE$\downarrow$
& MUSIQ$\uparrow$
& CLIP-IQA$\uparrow$
& LIQE-Mix$\uparrow$
& ARNIQA$\uparrow$
& TOPIQ$\uparrow$ \\
\midrule

\multicolumn{10}{c}{\textit{DBSR Models}} \\
\midrule

StableSR
& 24.2395 & 0.3129 & 0.5506 & 5.6054 & 58.4136
& 0.5627 & 3.4477 & 0.6580 & 0.4851 \\

SeeSR
& 23.6780 & 0.3194 & 0.6222 & 4.8095 & 68.6710
& 0.6935 & 4.0053 & 0.7301 & 0.6853 \\

SinSR
& 24.2883 & 0.3222 & 0.5414 & 5.8005 & 63.2651
& 0.6530 & 3.7998 & 0.6745 & 0.5796 \\

OSEDiff
& 23.7214 & 0.2942 & 0.6130 & 4.7203 & 67.9926
& 0.6611 & 4.2976 & 0.7030 & 0.6194 \\

AdcSR
& 23.3567 & 0.2845 & 0.6071 & 4.3420 & 68.0922
& 0.6846 & 4.3009 & 0.7215 & 0.6574 \\

\midrule
\multicolumn{10}{c}{\textit{DFSR Models}} \\
\midrule

SwinIR-L
& 23.9606 & 0.3152 & 0.5402 & 4.7492 & 60.1092
& 0.5246 & 3.5484 & 0.6607 & 0.5071 \\

\rowcolor{phoenixgray}
{\scriptsize (w/PhoenixSR)}
& 24.1746 & 0.2996 & 0.5675 & 4.3867 & 62.4468
& 0.5793 & 3.8263 & 0.6760 & 0.5372 \\

{\tiny Gain}
& \posgain{+0.2140}
& \posgain{-0.0156}
& \posgain{+0.0273}
& \posgain{-0.3625}
& \posgain{+2.3376}
& \posgain{+0.0547}
& \posgain{+0.2779}
& \posgain{+0.0153}
& \posgain{+0.0301} \\

\cmidrule{1-10}

HAT
& 24.7455 & 0.2931 & 0.5547 & 5.1686 & 58.2528
& 0.4622 & 3.4201 & 0.6467 & 0.4742 \\

\rowcolor{phoenixgray}
{\scriptsize (w/PhoenixSR)}
& 24.3991 & 0.2831 & 0.5853 & 4.6309 & 62.6893
& 0.5514 & 3.9650 & 0.6824 & 0.5446 \\

{\tiny Gain}
& \neggain{-0.3464}
& \posgain{-0.0100}
& \posgain{+0.0306}
& \posgain{-0.5377}
& \posgain{+4.4365}
& \posgain{+0.0892}
& \posgain{+0.5449}
& \posgain{+0.0357}
& \posgain{+0.0704} \\

\cmidrule{1-10}

Real-ESRGAN
& 24.2963 & 0.3116 & 0.5484 & 4.6792 & 61.0717
& 0.5274 & 3.6659 & 0.6738 & 0.5300 \\

\rowcolor{phoenixgray}
{\scriptsize (w/PhoenixSR)}
& 24.2379 & 0.3112 & 0.5437 & 4.3980 & 62.1329
& 0.5569 & 3.7818 & 0.6841 & 0.5378 \\

{\tiny Gain}
& \neggain{-0.0584}
& \posgain{-0.0004}
& \neggain{-0.0047}
& \posgain{-0.2812}
& \posgain{+1.0612}
& \posgain{+0.0295}
& \posgain{+0.1159}
& \posgain{+0.0103}
& \posgain{+0.0078} \\

\cmidrule{1-10}

SwinIR-M
& 23.9014 & 0.3235 & 0.5164 & 4.7827 & 58.1455
& 0.5170 & 3.4387 & 0.6513 & 0.4838 \\

\rowcolor{phoenixgray}
{\scriptsize (w/PhoenixSR)}
& 24.1663 & 0.3089 & 0.5477 & 4.4839 & 61.5357
& 0.5842 & 3.7218 & 0.6774 & 0.5226 \\

{\tiny Gain}
& \posgain{+0.2649}
& \posgain{-0.0146}
& \posgain{+0.0313}
& \posgain{-0.2988}
& \posgain{+3.3902}
& \posgain{+0.0672}
& \posgain{+0.2831}
& \posgain{+0.0261}
& \posgain{+0.0388} \\

\cmidrule{1-10}

SwinIR-S
& 23.9349 & 0.3764 & 0.3965 & 7.0025 & 46.0279
& 0.3890 & 2.0897 & 0.5831 & 0.3759 \\

\rowcolor{phoenixgray}
{\scriptsize (w/PhoenixSR)}
& 23.7942 & 0.3390 & 0.5176 & 4.2790 & 63.7584
& 0.6107 & 3.8624 & 0.6948 & 0.5674 \\

{\tiny Gain}
& \neggain{-0.1407}
& \posgain{-0.0374}
& \posgain{+0.1211}
& \posgain{-2.7235}
& \posgain{+17.7305}
& \posgain{+0.2217}
& \posgain{+1.7727}
& \posgain{+0.1117}
& \posgain{+0.1915} \\

\cmidrule{1-10}

SeeMoRe
& 24.8031 & 0.4907 & 0.2661 & 8.9587 & 37.7098
& 0.3919 & 1.5603 & 0.4023 & 0.2374 \\

\rowcolor{phoenixgray}
{\scriptsize (w/PhoenixSR)}
& 24.6089 & 0.3116 & 0.5164 & 4.2905 & 61.0420
& 0.5363 & 3.7589 & 0.6702 & 0.4916 \\

{\tiny Gain}
& \neggain{-0.1942}
& \posgain{-0.1791}
& \posgain{+0.2503}
& \posgain{-4.6682}
& \posgain{+23.3322}
& \posgain{+0.1444}
& \posgain{+2.1986}
& \posgain{+0.2679}
& \posgain{+0.2542} \\

\bottomrule
\end{tabular}

\vspace{-2mm}
\end{table*}

\subsection{Ablation Studies}
\label{sec:ablation}

We evaluate diffusion-transfer strategies and the contributions of PhoenixSR components. The tables report results from the full training schedule, whereas controlled comparisons and ablations use a fixed $80$K-iteration budget with the same number of student updates. Their absolute values are not directly comparable, as different training stages may correspond to different perception--distortion operating points. All ablation conclusions are drawn only within the same controlled setting.

\paragraph{Comparison with alternative transfer strategies.}
We compare PhoenixSR with strong task-only fine-tuning, output-level heterogeneous distillation, SDS, and DMD2-based distribution matching. All methods start from the same student checkpoint and use the same training data, number of student updates, paired SR supervision when applicable, and validation-based model selection protocol.

\begin{table*}[t]
\centering
\vspace{-1mm}
\caption{
Comparison of alternative diffusion-to-DFSR transfer strategies on RealSR.
}
\label{tab:transfer_comparison}
\scriptsize
\setlength{\tabcolsep}{4.2pt}
\renewcommand{\arraystretch}{0.90}

\begin{tabular}{l|ccc|ccc|ccc}
\toprule
& \multicolumn{3}{c|}{SwinIR-L}
& \multicolumn{3}{c|}{SwinIR-S}
& \multicolumn{3}{c}{SeeMoRe} \\
Method
& PSNR$\uparrow$ & LPIPS$\downarrow$ & MUSIQ$\uparrow$
& PSNR$\uparrow$ & LPIPS$\downarrow$ & MUSIQ$\uparrow$
& PSNR$\uparrow$ & LPIPS$\downarrow$ & MUSIQ$\uparrow$ \\
\midrule

Task-only strong FT
& 26.41 & 0.257 & 59.7
& 25.13 & 0.301 & 56.0
& 26.00 & 0.289 & 53.7 \\

Output KD + GT
& 25.97 & 0.274 & 56.7
& 24.77 & 0.327 & 55.8
& 25.98 & 0.301 & 52.1 \\

SDS + SFA
& 25.77 & 0.273 & 56.5
& 24.89 & 0.305 & 50.4
& 25.10 & 0.317 & 51.8 \\

DMD2 + SFA
& 25.55 & 0.256 & 57.8
& 24.78 & 0.311 & 54.7
& 24.87 & 0.304 & 53.6 \\

PhoenixSR w/o DRW
& 26.15 & 0.247 & 63.5
& 24.79 & 0.296 & 61.5
& 25.35 & 0.290 & 60.8 \\

\rowcolor{phoenixgray}
PhoenixSR
& 26.44 & 0.244 & 64.3
& 25.21 & 0.287 & 63.4
& 25.93 & 0.279 & 61.5 \\

\bottomrule
\end{tabular}

\vspace{-3mm}
\end{table*}

Table~\ref{tab:transfer_comparison} shows that PhoenixSR achieves a better fidelity--perception balance than task-only fine-tuning, output KD, SDS, and DMD2-based transfer across the three representative students. PhoenixSR without DRW already provides strong perceptual gains, while DRW mainly improves reconstruction fidelity and the overall balance. These results indicate that the main benefit comes from the proposed fidelity-aware distribution-level transfer rather than the auxiliary reweighting strategy. 
\vspace{-1mm}
\paragraph{Component ablation.}
We further analyze the main components on SwinIR-L. ``Score Adapt.'' denotes LoRA real-score adaptation together with REPA-regularized online fake-score optimization, while SFA denotes the sample-level fidelity anchor. Table~\ref{tab:ablation_main} shows that distribution matching without paired fidelity supervision causes severe degradation, confirming the importance of SFA for fidelity-sensitive SR. Combining SFA with score adaptation substantially improves perceptual quality, while DRW further improves PSNR and CLIP-IQA, yielding a better fidelity--perception balance. Additional ablations of score-model design, DRW, and the training schedule, together with robustness analyses, qualitative results, and the user study, are provided in the supplementary material.
\vspace{-2mm}

\begin{table}[H]
\centering
\vspace{-1mm}
\caption{Ablation of the main PhoenixSR components.}
\label{tab:ablation_main}
\scriptsize
\setlength{\tabcolsep}{3.2pt}
\renewcommand{\arraystretch}{0.90}
\resizebox{\linewidth}{!}{
\begin{tabular}{lccc|cccc}
\toprule
Variant & SFA & Score Adapt. & DRW
& PSNR$\uparrow$ & MANIQA$\uparrow$ & MUSIQ$\uparrow$ & CLIP-IQA$\uparrow$ \\
\midrule

Base SwinIR-L
& -- & -- & --
& 26.3071 & 0.5228 & 58.6388 & 0.4502 \\

SFA only
& \checkmark & -- & --
& 26.3124 & 0.5067 & 56.8245 & 0.4538 \\

w/o SFA
& -- & \checkmark & \checkmark
& 20.7107 & 0.3512 & 40.7941 & 0.3071 \\

SFA + Score Adapt.
& \checkmark & \checkmark & --
& 26.1587 & 0.5678 & 63.5729 & 0.4963 \\

\rowcolor[HTML]{EAEAEA}
Full PhoenixSR
& \checkmark & \checkmark & \checkmark
& 26.4409 & 0.5745 & 64.3102 & 0.5026 \\

\bottomrule
\end{tabular}
}
\vspace{-1mm}
\end{table}

\section{Conclusion}

We propose \textbf{PhoenixSR}, a generative heterogeneous distillation framework that transfers pretrained diffusion priors to independently designed diffusion-free SR networks through score-based distribution matching. Paired SR supervision preserves reconstruction fidelity, while HDA adapts score estimation to the SR setting and DRW further improves the fidelity--perception balance. Experiments across six feed-forward SR backbones and multiple real-world benchmarks demonstrate consistent perceptual improvements without changing the student inference architecture. Controlled comparisons with alternative transfer strategies further support the effectiveness of distribution-level diffusion-prior transfer for efficient SR.

\clearpage

\appendix

\begin{center}
    {\Large \textbf{Supplementary Material}}
\end{center}
\vspace{2mm}
\section{Extended Related Work}
\label{app:related_work}

\subsection{Real-World Image Super-Resolution}

Single-image super-resolution has progressed from CNN-based reconstruction models, including SRCNN~\citep{dong2014learning}, VDSR~\citep{kim2016accurate}, EDSR~\citep{lim2017enhanced}, and RCAN~\citep{zhang2018rcan}, to Transformer architectures such as SwinIR~\citep{liang2021swinir} and HAT~\citep{chen2023hat}. While these methods perform strongly under synthetic degradations, real-world SR additionally requires handling unknown blur, noise, compression, and camera-processing artifacts. Blind and real-world SR methods therefore estimate degradation kernels~\citep{gu2019blind}, construct practical degradation pipelines for synthesizing realistic LR--HR pairs~\citep{zhang2021designing,wang2021realesrgan,pengtowards}, or exploit task-specific transfer and raw-data supervision~\citep{yi2025fine,peng2024efficient,peng2024unveiling}. RealSR~\citep{cai2019toward} and DRealSR~\citep{wei2020component} provide real-captured evaluation settings.

Diffusion models provide stronger generative priors for perceptual restoration. SR3~\citep{saharia2022sr3} formulates SR as iterative denoising, while latent diffusion~\citep{rombach2022high} enables scalable pretrained generative priors. StableSR~\citep{wang2023stablesr}, ResShift~\citep{yue2024resshift}, DiffBIR~\citep{lin2024diffbir}, SeeSR~\citep{wu2024seesr}, PASD~\citep{yang2024pixel}, and SUPIR~\citep{yu2024scaling} exploit such priors for real-world restoration. SinSR~\citep{wang2024sinsr}, OSEDiff~\citep{wu2024osediff}, AdcSR~\citep{chen2025adcsr}, GDPO-SR~\citep{yi2026gdpo}, OP4KSR~\citep{deng2026op4ksr}, and trajectory-consistency learning~\citep{deng2026joint} pursue efficient or one-step generation, but still use generation pipelines designed around diffusion priors.

DM-SR~\citep{park2026dmsr} further introduces distribution matching into SR by mapping degraded inputs toward distributions compatible with a pretrained diffusion model, while the diffusion denoiser remains part of inference. PhoenixSR instead uses the pretrained diffusion model only as training-time distributional supervision and removes all diffusion components after distillation.

\subsection{Lightweight and Efficient Super-Resolution}

Efficient SR is important for mobile, embedded, and real-time applications. Early methods reduce computation by operating mainly in LR feature space and using efficient upsampling~\citep{shi2016real,dong2016accelerating}. CARN~\citep{ahn2018fast} improves feature reuse through cascading residual connections, while IDN~\citep{hui2018fast} and IMDN~\citep{hui2019lightweight} introduce information-distillation mechanisms for compact feature representations. RLFN~\citep{kong2022rlfn}, LAPAR~\citep{li2020lapar}, and SAFMN~\citep{sun2023safmn} further improve the efficiency--accuracy trade-off through lightweight feature aggregation, pixel-adaptive regression, or spatial feature modulation.

Transformer-based restoration models provide another family of feed-forward SR networks. SwinIR~\citep{liang2021swinir} introduces shifted-window attention for image restoration, while HAT~\citep{chen2023hat} combines window and channel attention for stronger reconstruction capability. SeeMoRe~\citep{zamfir2024seemore} achieves efficient detail restoration with approximately one million parameters through expert-based feature modeling. Recent alternatives use hierarchical state-space modeling~\citep{deng2026ihmambasr}, query state-space models for burst SR~\citep{di2025qmambabsr}, or explicit 2D Gaussian representations for spatial and spatio-temporal continuous SR~\citep{peng2025pixel,shi2026gsstvsrultraefficientcontinuousspatiotemporal}.

A closely related direction is heterogeneous knowledge transfer from diffusion models to efficient SR networks. DTKD~\citep{park2026dtkd} transfers perceptual knowledge from a diffusion SR teacher to a Transformer student through frequency-aware output distillation. PhoenixSR shares the goal of diffusion-to-feed-forward knowledge transfer, but transfers the pretrained diffusion prior through score-based distribution matching rather than directly matching sampled teacher outputs.

\subsection{Broader Image Restoration and Video Enhancement}

Ideas for efficient representation and texture modeling also arise in neighboring low-level vision tasks. Adaptive feature de-drifting improves robustness under image compression~\citep{peng2024lightweight}, while texture-aware state-space modeling targets complex restoration patterns~\citep{peng2025directing}. Deraining research spans structure-preserving priors, multi-scale Transformers, efficient ultra-high-definition processing, and synergistic convolution~\citep{yi2021structure,chen2024rethinking,11408230,peng2025boosting}. Ultra-high-definition restoration increasingly emphasizes efficient global, high-frequency, and cluster-aware computation~\citep{chen2025mixnet,wu2025ultra,wu2026scan}. Related degradation-aware designs address daytime and nighttime dehazing~\citep{yi2021efficient,liu2023multi,liu2023nighthazeformer,wu2026efficient}, exposure correction~\citep{wang2023decoupling}, all-in-one restoration~\citep{Perceive-IR,ClearAIR}, and underwater restoration~\citep{UniUIR}. For video, progressive multi-frame quantization improves enhancement efficiency~\citep{fengpmq}, continuous video SR can be formulated with 2D Gaussian splatting~\citep{shi2026gsstvsrultraefficientcontinuousspatiotemporal}, and large-scale video diffusion models provide increasingly capable generative priors~\citep{wu2025hunyuanvideo}. These developments motivate restoration methods that combine strong learned priors with efficient deployment.

\subsection{Knowledge Distillation and Diffusion Distillation}

Knowledge distillation transfers knowledge from a high-capacity teacher to a smaller student~\citep{hinton2015distilling,gou2021knowledge}. Beyond response-level supervision, later methods align intermediate features~\citep{romero2015fitnets,he2019knowledge}, similarity structures~\citep{tung2019similarity}, and relational information~\citep{park2019relational}. Such approaches are most natural when teacher and student possess compatible representations or deterministic output correspondence.

Diffusion distillation primarily accelerates multi-step diffusion generators. Progressive distillation~\citep{salimans2022progressive}, consistency models~\citep{song2023consistency}, and guided diffusion distillation~\citep{meng2023distillation} reduce the number of denoising steps while retaining diffusion-derived generators. Distribution Matching Distillation (DMD)~\citep{yin2024one} combines distribution matching through real--fake score differences with a teacher-output regression loss for stable training. DMD2~\citep{yin2024improved} removes the regression requirement using two-time-scale updates and additionally introduces real-data adversarial supervision and a training procedure for few-step generation. Related score-distillation approaches such as SDS~\citep{poole2023dreamfusion} and VSD~\citep{wang2023prolificdreamer} also demonstrate the effectiveness of pretrained diffusion scores as distribution-level supervision.

DMD has also been extended to video SR. DUO-VSR~\citep{lv2026duovsr} combines distribution matching with additional supervision to obtain one-step video SR, but its generator remains diffusion-derived. In contrast, PhoenixSR transfers a pretrained diffusion prior to independently designed diffusion-free SR students and preserves their original feed-forward inference architectures.

\subsection{Positioning with Closely Related Diffusion-Transfer Methods}
\label{app:related_positioning}

Several recent works are closely related to PhoenixSR in their use of diffusion priors for efficient generation or super-resolution. We therefore further clarify the specific setting addressed by PhoenixSR and its differences from DTKD~\citep{park2026dtkd}, DM-SR~\citep{park2026dmsr}, RSD~\citep{selikhanovych2026rsd}, TSD-SR~\citep{dong2025tsdsr}, and Cross-Space Distillation~\citep{nguyen2026crossspace}.

The novelty of PhoenixSR does not simply lie in transferring knowledge from a diffusion model to an efficient SR network. DTKD has already demonstrated diffusion-to-Transformer heterogeneous distillation while retaining feed-forward student inference. Instead, PhoenixSR focuses on transferring a pretrained diffusion prior through \emph{score-based distribution matching} into pre-existing, independently designed diffusion-free SR architectures, without retaining a diffusion-derived generator or modifying the native student inference graph.

\begin{table*}[t]
\centering
\caption{
\textbf{Positioning of PhoenixSR with closely related diffusion-transfer methods.}
``Native DFSR'' denotes a conventional feed-forward SR architecture designed independently from the diffusion teacher. ``Diffusion-derived inference'' indicates whether the deployed model still relies on a diffusion model or a diffusion-derived generator.
}
\label{tab:app_related_positioning}
\resizebox{\textwidth}{!}{
\begin{tabular}{lccccc}
\toprule
Method
& Task
& Transfer mechanism
& Deployed model
& Native DFSR
& Diffusion-derived inference \\
\midrule

DTKD~\citep{park2026dtkd}
& SR
& Frequency-aware output KD
& Transformer SR student
& \checkmark
& No \\

DM-SR~\citep{park2026dmsr}
& SR
& Distribution alignment
& Pretrained diffusion model with learned input mapping
& --
& Yes \\

Cross-Space Distillation~\citep{nguyen2026crossspace}
& Image generation
& Distribution-level diffusion distillation
& One-step diffusion student
& --
& Yes \\

\rowcolor{phoenixgray}
PhoenixSR
& SR
& Score-based distribution matching
& CNN / Transformer / lightweight SR students
& \checkmark
& No \\

\bottomrule
\end{tabular}
}
\end{table*}

\paragraph{Difference from DTKD.}
DTKD~\citep{park2026dtkd} transfers perceptual knowledge from a diffusion SR teacher to a Transformer student through output-level supervision. It decomposes the output discrepancy into frequency sub-bands and emphasizes selected frequency components during distillation. PhoenixSR shares the broader goal of diffusion-to-feed-forward SR transfer, but adopts a different form of supervision. Rather than requiring the student to imitate a sampled diffusion output, PhoenixSR uses the pretrained diffusion model to provide distribution-level guidance through real--fake score differences. This avoids enforcing direct sample-wise correspondence with a particular diffusion reconstruction, which may contain perceptually plausible details that do not exactly agree with the paired HR target.

PhoenixSR is also evaluated across CNN, Transformer, and lightweight SR architectures. These students are independently designed feed-forward SR networks rather than architectures derived from the diffusion teacher, and their original inference graphs remain unchanged after distillation.

\paragraph{Difference from DM-SR.}
DM-SR~\citep{park2026dmsr} also introduces distribution matching into super-resolution, but addresses a different problem. It learns an encoder that maps degraded LR observations toward a distribution compatible with a pretrained diffusion model, after which the pretrained diffusion model remains responsible for restoration at inference time.

PhoenixSR instead performs distribution matching directly on an independently designed diffusion-free SR student. The diffusion model is used only to provide training-time supervision. After distillation, the diffusion score models, VAE, and other auxiliary components are removed, and the deployed model is exactly the original feed-forward SR student. Therefore, PhoenixSR transfers the diffusion prior into the student itself rather than preserving the diffusion model as part of the restoration pipeline.

\paragraph{Difference from RSD and TSD-SR.}
RSD~\citep{selikhanovych2026rsd} combines distribution-based ResShift distillation with ground-truth perceptual supervision and an adversarial loss. TSD-SR~\citep{dong2025tsdsr} incorporates real-image reference scores and reconstruction supervision into one-step diffusion SR. Both retain diffusion-derived generators at inference and already combine diffusion guidance with restoration supervision. Building on this line of work, PhoenixSR develops HDA and DRW for transfer into pre-existing, independently designed diffusion-free SR networks. Its contribution lies in adapting score estimation and guidance to this heterogeneous setting while preserving the student's original inference graph.

\paragraph{Difference from Cross-Space Distillation.}
Cross-Space Distillation~\citep{nguyen2026crossspace} studies distribution-level knowledge transfer between diffusion teachers and one-step diffusion students whose latent resolutions or VAE spaces differ. Its heterogeneous setting therefore arises from incompatibility between different diffusion latent spaces, while the deployed student remains a diffusion-derived generator.

PhoenixSR considers a different form of heterogeneity: the student is not a diffusion model, but a native deterministic SR network such as SwinIR, HAT, Real-ESRGAN, or SeeMoRe. The diffusion latent space is used only as a supervision space during training and does not become part of the student architecture or inference process.

Taken together, PhoenixSR specifically studies \emph{score-based distribution-level transfer of a pretrained diffusion prior into pre-existing, independently designed diffusion-free SR architectures while preserving their native feed-forward inference graphs}. HDA enables this distribution-level transfer to operate under the paired and fidelity-sensitive constraints of SR, while DRW further addresses sample-wise variation in the stability of online fake-score guidance.

\paragraph{Discussion of direct quantitative comparison.}
A strictly controlled comparison between heterogeneous distillation methods requires the same student initialization, training data, degradation pipeline, optimization budget, paired supervision, and evaluation protocol. This is particularly important in real-world SR, where changes in degradation synthesis and reconstruction objectives can substantially affect both distortion and perceptual metrics.

For DTKD~\citep{park2026dtkd}, the published experiments use a diffusion SR teacher and a SwinIR-based student under their own training and distillation setting. At the time of submission, we could not identify an official public implementation or pretrained distillation checkpoint that would allow DTKD to be reproduced under the controlled PhoenixSR setting. Directly inserting its reported numbers into our quantitative tables would therefore mix different training protocols and would not constitute a controlled comparison. Instead, the main paper includes an output-level heterogeneous KD baseline trained with the same student initialization, training data, update budget, paired supervision, and model-selection protocol as PhoenixSR. This baseline is intended to isolate the difference between direct teacher-output supervision and distribution-level transfer, rather than to serve as a reproduction of DTKD.

Similarly, at the time of submission, we could not identify an official public implementation of DM-SR~\citep{park2026dmsr} that could be evaluated under our unified protocol. Moreover, DM-SR and PhoenixSR target different deployment settings: DM-SR retains the pretrained diffusion model in its restoration pipeline, whereas PhoenixSR removes all diffusion-related components after training. We therefore include representative diffusion-based SR methods as inference-time reference baselines and use controlled SDS/DMD-based variants to analyze distribution-level transfer within the native DFSR setting.

These comparisons separate two related but distinct questions. Prior work has established that diffusion priors can improve perceptual SR and can be transferred to efficient students. PhoenixSR instead investigates whether a pretrained diffusion prior can be transferred through score-based distribution matching directly into independently designed native diffusion-free SR architectures, while preserving their original feed-forward inference path.

\section{Additional Method Details}
\label{app:method_details}

\subsection{Latent-Space Distribution Matching}

PhoenixSR uses Stable Diffusion~2.1 as a training-only diffusion prior. Let the diffusion-free SR student produce
\begin{equation}
    \hat{\mathbf{x}}_0 = G_\theta(\mathbf{y}),
\end{equation}
and let $E$ denote the frozen Stable Diffusion VAE encoder. The generated image is mapped to the latent space as
\begin{equation}
    \hat{\mathbf{z}}_0 = E(\hat{\mathbf{x}}_0),
    \qquad
    \hat{\mathbf{z}}_t
    =
    \alpha_t\hat{\mathbf{z}}_0
    +
    \sigma_t\boldsymbol{\epsilon},
    \quad
    \boldsymbol{\epsilon}\sim\mathcal{N}(\mathbf{0},\mathbf{I}).
    \label{eq:app_latent_forward}
\end{equation}

The real-score model and fake-score model predict noise on the same noisy generated latent:
\begin{equation}
    \boldsymbol{\epsilon}_{\mathrm{real}}
    =
    \boldsymbol{\epsilon}_{\mathrm{real}}
    (\hat{\mathbf{z}}_t,t,\mathbf{c}),
    \qquad
    \boldsymbol{\epsilon}_{\mathrm{fake}}
    =
    \boldsymbol{\epsilon}_{\mathrm{fake}}
    (\hat{\mathbf{z}}_t,t,\mathbf{c}),
\end{equation}
where $\mathbf{c}$ is the text condition. In the reported experiments, all samples use the same fixed quality prompt.

Under noise prediction, the diffusion score is proportional to the negative predicted noise. Thus, the real--fake score discrepancy provides distribution-level supervision without requiring the SR student to share the diffusion architecture.

\subsection{Denoised-Latent DMD Implementation}
\label{app:dmd_impl}

For stable optimization, our implementation realizes the score discrepancy through an equivalent denoised-latent parameterization. Given cumulative diffusion coefficient $\bar{\alpha}_t$, define
\begin{equation}
    a_t=\sqrt{\bar{\alpha}_t},
    \qquad
    b_t=\sqrt{1-\bar{\alpha}_t}.
\end{equation}
The real and fake predictions of the clean latent are
\begin{equation}
    \tilde{\mathbf{z}}_{0}^{\,\mathrm{real}}
    =
    \frac{
    \hat{\mathbf{z}}_t
    -
    b_t\boldsymbol{\epsilon}_{\mathrm{real}}
    }{a_t},
    \qquad
    \tilde{\mathbf{z}}_{0}^{\,\mathrm{fake}}
    =
    \frac{
    \hat{\mathbf{z}}_t
    -
    b_t\boldsymbol{\epsilon}_{\mathrm{fake}}
    }{a_t}.
    \label{eq:app_denoised_latent}
\end{equation}

We compute the per-sample distributional guidance
\begin{equation}
    \mathbf{g}_i
    =
    \frac{
    \tilde{\mathbf{z}}_{0,i}^{\,\mathrm{fake}}
    -
    \tilde{\mathbf{z}}_{0,i}^{\,\mathrm{real}}
    }{
    \operatorname{mean}
    \left(
    \left|
    \hat{\mathbf{z}}_{0,i}
    -
    \tilde{\mathbf{z}}_{0,i}^{\,\mathrm{real}}
    \right|
    \right)
    +\epsilon
    }.
    \label{eq:app_dmd_guidance}
\end{equation}
The denominator prevents low-noise samples with very small denoising distances from dominating optimization. For additional stability, the root-mean-square magnitude of $\mathbf{g}_i$ is clipped when it exceeds the configured threshold.

A detached surrogate target is then formed as
\begin{equation}
    \mathbf{z}^{\,\mathrm{tar}}_{0,i}
    =
    \operatorname{sg}
    \left(
    \hat{\mathbf{z}}_{0,i}-\mathbf{g}_i
    \right),
\end{equation}
where $\operatorname{sg}(\cdot)$ denotes stop-gradient. The corresponding per-sample DMD surrogate is
\begin{equation}
    \mathcal{L}_{\mathrm{DMD}}^{i}
    =
    \frac{1}{2}
    \left\|
    \hat{\mathbf{z}}_{0,i}
    -
    \mathbf{z}^{\,\mathrm{tar}}_{0,i}
    \right\|_2^2.
    \label{eq:app_dmd_surrogate}
\end{equation}
Because the target is detached, differentiating Eq.~\eqref{eq:app_dmd_surrogate} produces the desired distributional guidance without back-propagating through the score models.

\subsection{Detailed Heterogeneous Distribution Adaptation}

\paragraph{LoRA-adapted real-score model.}
The pretrained Stable Diffusion U-Net is used as the real-score model. Its original parameters remain frozen, and LoRA~\citep{hu2022lora} adapters are inserted into the attention and projection layers. For an adapted linear layer,
\begin{equation}
    \mathbf{W}
    =
    \mathbf{W}_0
    +
    \frac{\gamma}{r}\mathbf{B}\mathbf{A},
    \label{eq:app_lora}
\end{equation}
where $\mathbf{W}_0$ is frozen, $\mathbf{A}$ and $\mathbf{B}$ are trainable low-rank matrices, $r$ is the rank, and $\gamma$ is the LoRA scaling factor.

\begin{figure}[t]
    \centering
    \includegraphics[width=\linewidth]{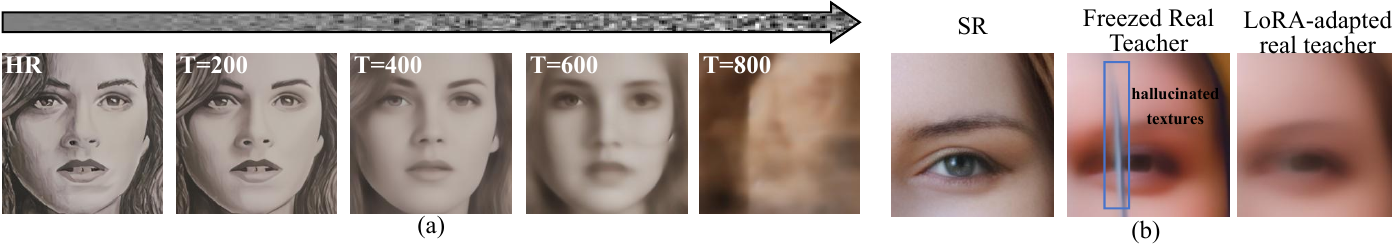}
    \caption{
    \textbf{Analysis of diffusion guidance for SR distillation.}
    (a) \textbf{Effect of timestep selection.} Low-noise states preserve more input-consistent structures, while stronger noise makes diffusion guidance increasingly dominated by the generative prior and may introduce unfaithful details.
    (b) \textbf{Visualization of real-score adaptation.} The frozen real-score model may convert restoration artifacts into plausible but unfaithful textures, while LoRA adaptation alleviates such artifact-to-texture degeneration and provides more SR-compatible guidance.
    }
    \label{fig:app_guidance_analysis}
\end{figure}

Given a paired HR image $\mathbf{x}$, its latent representation is
\begin{equation}
    \mathbf{z}^{HR}_0=E(\mathbf{x}),
    \qquad
    \mathbf{z}^{HR}_t
    =
    \alpha_t\mathbf{z}^{HR}_0
    +
    \sigma_t\boldsymbol{\epsilon}.
\end{equation}
Only the LoRA parameters are optimized using
\begin{equation}
    \mathcal{L}_{\mathrm{real}}
    =
    \mathbb{E}
    \left[
    \left\|
    \boldsymbol{\epsilon}_{\mathrm{real}}
    (\mathbf{z}^{HR}_t,t,\mathbf{c})
    -
    \boldsymbol{\epsilon}
    \right\|_2^2
    \right].
    \label{eq:app_real_loss}
\end{equation}
This approach limits adaptation while retaining the pretrained backbone. As illustrated in Fig.~\ref{fig:app_guidance_analysis}(b), the frozen real-score model may transform restoration artifacts into perceptually plausible but unfaithful textures, whereas LoRA adaptation alleviates this behavior and provides guidance that is better aligned with SR reconstruction.

\paragraph{Representation-regularized fake-score model.}
The fake-score model is initialized by copying the pretrained diffusion U-Net and is fully optimized online to track the evolving student distribution:
\begin{equation}
    \mathcal{L}_{\mathrm{fake}}
    =
    \mathbb{E}
    \left[
    \left\|
    \boldsymbol{\epsilon}_{\mathrm{fake}}
    (\hat{\mathbf{z}}_t,t,\mathbf{c})
    -
    \boldsymbol{\epsilon}
    \right\|_2^2
    \right].
\end{equation}

Because the student distribution changes continuously, the fake-score model may otherwise overfit transient artifacts. We therefore introduce representation alignment following REPA~\citep{yu2025representation}. Let $\mathbf{f}^{l}$ denote an intermediate fake-UNet feature, $\mathcal{P}$ a trainable projection, and $\mathbf{h}_{\mathrm{DINO}}$ the representation produced by a frozen DINOv3 encoder~\citep{simeoni2025dinov3}. The REPA objective is
\begin{equation}
\mathcal{L}_{\mathrm{REPA}}
=
-\mathbb{E}
\left[
\frac{1}{N_p}
\sum_{n=1}^{N_p}
\operatorname{sim}
\left(
\mathbf{u}^{l}_{n},
\operatorname{sg}(\mathbf{v}_{n})
\right)
\right],
\label{eq:app_repa}
\end{equation}
where
\begin{equation}
    \mathbf{u}^{l}=\mathcal{P}(\mathbf{f}^{l}),
    \qquad
    \mathbf{v}=\mathbf{h}_{\mathrm{DINO}}(\hat{\mathbf{x}}_0),
\end{equation}
and $\operatorname{sim}$ denotes cosine similarity. The complete fake-score objective is
\begin{equation}
    \mathcal{L}_{\mathrm{fake}}^{\mathrm{total}}
    =
    \mathcal{L}_{\mathrm{fake}}
    +
    \lambda_{\mathrm{repa}}
    \mathcal{L}_{\mathrm{REPA}}.
\end{equation}

\paragraph{Sample-level fidelity anchor.}
Distribution matching alone does not guarantee correspondence with the paired HR target. We therefore constrain the SR generator using
\begin{equation}
    \mathcal{L}_{\mathrm{SFA}}
    =
    \lambda_{\mathrm{pix}}\mathcal{L}_{\mathrm{pix}}
    +
    \lambda_{\mathrm{per}}\mathcal{L}_{\mathrm{per}}
    +
    \lambda_{\mathrm{adv}}\mathcal{L}_{\mathrm{adv}},
    \label{eq:app_sfa}
\end{equation}
with
\begin{equation}
    \mathcal{L}_{\mathrm{pix}}
    =
    \mathbb{E}
    \left[
    \left\|
    G_\theta(\mathbf{y})-\mathbf{x}
    \right\|_1
    \right],
\end{equation}
and
\begin{equation}
    \mathcal{L}_{\mathrm{per}}
    =
    \mathbb{E}
    \left[
    \left\|
    \varphi(G_\theta(\mathbf{y}))
    -
    \varphi(\mathbf{x})
    \right\|_1
    \right],
\end{equation}
where $\varphi$ is the feature extractor used for perceptual supervision~\citep{johnson2016perceptual}. The adversarial component uses the same generator/discriminator objective as the underlying real-world SR training framework.

The student generator is therefore optimized jointly by sample-level fidelity supervision and distribution-level diffusion guidance:
\begin{equation}
    \nabla_\theta\mathcal{L}_{G}
    =
    \lambda_{\mathrm{dmd}}
    \nabla_\theta\mathcal{L}_{\mathrm{DMD}}
    +
    \nabla_\theta\mathcal{L}_{\mathrm{SFA}}.
\end{equation}

\subsection{Detailed Directional Reliability Weighting}
\label{app:drw_details}

DRW estimates whether the online fake-score model behaves consistently around the generated and real-data neighborhoods. For the $i$-th paired sample, let
\begin{equation}
    \mathbf{z}^{G,i}_0
    =
    E(G_\theta(\mathbf{y}^{i})),
    \qquad
    \mathbf{z}^{HR,i}_0
    =
    E(\mathbf{x}^{i}).
\end{equation}
The same timestep and Gaussian noise are applied to both:
\begin{equation}
    \mathbf{z}^{G,i}_t
    =
    \alpha_t\mathbf{z}^{G,i}_0
    +
    \sigma_t\boldsymbol{\epsilon}^{i},
    \qquad
    \mathbf{z}^{HR,i}_t
    =
    \alpha_t\mathbf{z}^{HR,i}_0
    +
    \sigma_t\boldsymbol{\epsilon}^{i}.
\end{equation}

Their fake-score denoising residuals are
\begin{equation}
    \mathbf{r}_{G}^{i}
    =
    \boldsymbol{\epsilon}_{\mathrm{fake}}
    (\mathbf{z}^{G,i}_t,t,\mathbf{c}^{i})
    -
    \boldsymbol{\epsilon}^{i},
\end{equation}
and
\begin{equation}
    \mathbf{r}_{HR}^{i}
    =
    \boldsymbol{\epsilon}_{\mathrm{fake}}
    (\mathbf{z}^{HR,i}_t,t,\mathbf{c}^{i})
    -
    \boldsymbol{\epsilon}^{i}.
\end{equation}
We compute
\begin{equation}
    s_i
    =
    \frac{
    \langle
    \mathbf{r}_{G}^{i},
    \mathbf{r}_{HR}^{i}
    \rangle
    }{
    \|\mathbf{r}_{G}^{i}\|_2
    \|\mathbf{r}_{HR}^{i}\|_2
    }.
    \label{eq:app_drw_score}
\end{equation}

Importantly, $s_i$ is not an image-quality score. It measures the directional consistency of the fake-score model around generated and real-data samples. Higher consistency indicates that the fake-score model exhibits more similar local denoising behavior around the two regions.

The reliability weight is
\begin{equation}
    \omega_i
    =
    \operatorname{clip}
    \left(
    N
    \frac{
    \exp(s_i/\tau)
    }{
    \sum_{j=1}^{N}\exp(s_j/\tau)
    },
    \omega_{\min},
    \omega_{\max}
    \right).
    \label{eq:app_drw_weight}
\end{equation}
The final weighted DMD loss is
\begin{equation}
    \mathcal{L}_{\mathrm{DMD}}^{\mathrm{DRW}}
    =
    \frac{1}{N}
    \sum_{i=1}^{N}
    \operatorname{sg}(\omega_i)
    \mathcal{L}_{\mathrm{DMD}}^{i}.
\end{equation}
During the generator update, no gradient is propagated through $s_i$ or $\omega_i$.

\paragraph{Distributed normalization.}
Training uses one image per GPU. Therefore, computing the softmax independently on each GPU would degenerate to $\omega_i=1$. Instead, the directional similarities are gathered across distributed workers before the softmax. With eight GPUs and one sample per GPU, the normalization pool contains eight samples. Only the resulting scalar weights are used to modulate the per-sample DMD contributions.

DRW does not change the real--fake score direction of an individual sample. It only changes that sample's relative contribution to the aggregated update.

\section{Training Strategy and Implementation Details}
\label{app:training_details}

\subsection{Timestep Truncation and Noise Shift}

The diffusion timestep controls the strength of the generative prior involved in distribution matching. For fidelity-sensitive SR, excessively noisy states may weaken the correspondence with the student reconstruction and make the guidance increasingly prior-driven. As shown in Fig.~\ref{fig:app_guidance_analysis}(a), low-noise states preserve more input-consistent structures, whereas stronger noise leads to increasingly generative predictions and may introduce unfaithful details. We therefore restrict DMD supervision to
\begin{equation}
    t\in[0,200).
\end{equation}

Within this range, the implementation applies a monotonic noise-shift transform. A base timestep $t_b$ is sampled uniformly and normalized as
\begin{equation}
    u
    =
    \frac{t_b-t_{\min}}
    {t_{\max}-t_{\min}-1}.
\end{equation}
The shifted normalized timestep is
\begin{equation}
    \phi_s(u)
    =
    \frac{s\,u}
    {1+(s-1)u},
    \label{eq:app_noise_shift}
\end{equation}
and the final discrete timestep is obtained by mapping $\phi_s(u)$ back to $[t_{\min},t_{\max})$ and rounding to the nearest integer.

The shift parameter $s$ is changed between training stages, allowing the distribution of sampled noise levels to vary during optimization while remaining within the fidelity-oriented range $[0,200)$.
\subsection{Warm-Up and Asynchronous Optimization}

The SR generator and the two score models evolve at different rates. Applying DMD before the fake-score model has adapted to the current student distribution can produce unstable guidance. We therefore first warm up the score models while keeping generator updates disabled.

Following the two-time-scale training principle of DMD2~\citep{yin2024improved}, score models are refreshed more frequently than the SR generator after warm-up. The real-score LoRA is optimized on noisy HR latents, while the fake-score model is optimized on detached noisy student latents. The generator is updated only at its configured interval using the latest real/fake score estimates, DRW weights, and SFA supervision. The discriminator follows its own adversarial update schedule.

This asynchronous optimization reduces stale fake-score estimates and separates score-model fitting from generator optimization.

\subsection{Complete Training Procedure}

Algorithm~\ref{alg:app_phoenix} summarizes the complete PhoenixSR training procedure.

\begin{algorithm}[t]
\caption{PhoenixSR Distillation}
\label{alg:app_phoenix}
\begin{algorithmic}[1]
\REQUIRE Paired data $\mathcal{D}$; DFSR student $G_\theta$; VAE encoder $E$; real-score model; fake-score model; DINOv3 encoder; discriminator.
\STATE Initialize $G_\theta$ from the corresponding DFSR checkpoint.
\STATE Initialize the real-score and fake-score models from Stable Diffusion~2.1.
\STATE Freeze the real-score backbone and optimize only its LoRA adapters.
\STATE Fully optimize the fake-score model with denoising and REPA objectives.
\FOR{training iteration $k$}
    \STATE Sample $(\mathbf{y},\mathbf{x})\sim\mathcal{D}$.
    \STATE Compute $\hat{\mathbf{x}}_0=G_\theta(\mathbf{y})$.
    \STATE Encode $\hat{\mathbf{z}}_0=E(\hat{\mathbf{x}}_0)$ and $\mathbf{z}^{HR}_0=E(\mathbf{x})$.
    \STATE Sample $t\in[0,200)$ and apply the noise-shift transform.
    \STATE Update the real-score LoRA using noisy HR latents.
    \STATE Update the fake-score model using detached noisy student latents and REPA.
    \IF{generator update is enabled after warm-up}
        \STATE Evaluate real/fake predictions on the same noisy generated latent.
        \STATE Compute per-sample DMD guidance using Eq.~\eqref{eq:app_dmd_guidance}.
        \STATE Compute DRW weights using Eqs.~\eqref{eq:app_drw_score}--\eqref{eq:app_drw_weight}.
        \STATE Compute the sample-level fidelity anchor in Eq.~\eqref{eq:app_sfa}.
        \STATE Update $G_\theta$ with weighted DMD and SFA supervision.
    \ENDIF
    \STATE Update the discriminator according to the adversarial schedule.
\ENDFOR
\STATE Discard VAE, diffusion score models, LoRA adapters, DINO/REPA modules, and discriminator.
\STATE Deploy $G_\theta$ only.
\end{algorithmic}
\end{algorithm}

\subsection{Experimental Configuration}

The HR training set combines DIV2K~\citep{agustsson2017ntire}, Flickr2K~\citep{lim2017enhanced}, OST~\citep{wang2018recovering}, WED~\citep{ma2017waterloo}, FFHQ~\citep{karras2019style}, Manga109~\citep{matsui2017sketch}, and SCUT-CTW1500~\citep{liu2019curved}. LR inputs are synthesized using the Real-ESRGAN degradation pipeline~\citep{wang2021realesrgan}. Training crops are $256\times256$ in HR and $64\times64$ in LR for $4\times$ SR. Experiments are conducted on eight NVIDIA V100 GPUs with one sample per GPU.

Unless otherwise specified, we use the following default configuration. The SR generator is optimized with Adam~\citep{kingma2015adam} using a learning rate of $3\times10^{-5}$ and $(\beta_1,\beta_2)=(0.9,0.99)$. The real-score U-Net uses LoRA with rank $16$, scaling factor $32$, and dropout $0$, while the pretrained diffusion backbone remains frozen. The fake-score U-Net is initialized from the pretrained diffusion U-Net and is fully optimized online. Both score models are optimized with AdamW~\citep{loshchilov2019decoupled} using a learning rate of $3\times10^{-5}$.

For the sample-level fidelity anchor, we set
$\lambda_{\mathrm{pix}}=0.1$,
$\lambda_{\mathrm{per}}=0.01$, and
$\lambda_{\mathrm{adv}}=0.1$.
The distribution-matching weight is set to
$\lambda_{\mathrm{dmd}}=1.0$, and the REPA regularization weight is
$\lambda_{\mathrm{repa}}=0.1$.
For DRW, the temperature is set to $\tau=0.2$, and the sample weights are clipped to $[0.2,3.0]$. DMD supervision uses the truncated timestep range $[0,200)$ together with the noise-shift strategy described above.

The reported experiments use the same fixed text prompt for all samples:
\begin{quote}
\small
``masterpiece, best quality, ultra-detailed, high resolution, highly detailed, sharp focus, extremely intricate, professional photography, high-definition.''
\end{quote}

\paragraph{Training cost.}
The training cost varies across student backbones because of their different model capacities and convergence behaviors. In our experiments, PhoenixSR distillation typically requires approximately $3$--$6$ days on eight NVIDIA V100 GPUs, with training lengths ranging from roughly $10^5$ to $10^6$ iterations. The additional cost is confined to the offline distillation stage, during which the diffusion score models, VAE encoder, DINO/REPA modules, and discriminator are used only to provide training supervision. After distillation, all auxiliary modules are discarded, and only the original feed-forward SR student is retained for inference. Therefore, PhoenixSR introduces no additional model parameters, FLOPs, or inference latency at deployment.

\section{Additional Ablation Studies}
\label{app:ablations}
\suppressfloats[t]

Unless otherwise specified, the ablation experiments use SwinIR-L under the same controlled $80$K-iteration setting as the main component ablation.

\subsection{Robustness Across Random Seeds}
\label{app:seed_robustness}

Since distribution-matching objectives involve stochastic diffusion perturbations and online score-model optimization, we additionally evaluate the robustness of PhoenixSR across independent training runs. We repeat the distillation experiments with three random seeds for one representative large student (SwinIR-L) and one lightweight student (SeeMoRe) on RealSR. The pretrained base checkpoints are deterministic references and are therefore evaluated once, while the distillation variants are independently trained for each seed. We report the mean and standard deviation over the three runs.

\begin{table*}[t]
\centering
\caption{\textbf{Robustness across random seeds on RealSR.}
PhoenixSR variants are independently trained with three random seeds and reported as mean$\pm$standard deviation. The base checkpoints are fixed references evaluated once.}
\label{tab:app_seed_robustness}
\resizebox{\textwidth}{!}{
\begin{tabular}{llccccc}
\toprule
Student & Method
& PSNR$\uparrow$
& LPIPS$\downarrow$
& MANIQA$\uparrow$
& MUSIQ$\uparrow$
& CLIP-IQA$\uparrow$ \\
\midrule
\multirow{3}{*}{SwinIR-L}
& Base checkpoint
& 26.30 & 0.253 & 0.522 & 58.6 & 0.45 \\
& PhoenixSR w/o DRW
& 26.20$\pm$0.03 & 0.251$\pm$0.004 & 0.557$\pm$0.007 & 62.9$\pm$0.3 & 0.481$\pm$0.005 \\
& PhoenixSR
& 26.40$\pm$0.04 & 0.247$\pm$0.008 & 0.568$\pm$0.006 & 64.1$\pm$0.5 & 0.493$\pm$0.007 \\
\midrule
\multirow{3}{*}{SeeMoRe}
& Base checkpoint
& 26.0 & 0.295 & 0.347 & 48.7 & 0.381 \\
& PhoenixSR w/o DRW
& 25.41$\pm$0.07 & 0.277$\pm$0.009 & 0.511$\pm$0.008 & 60.9$\pm$0.4 & 0.519$\pm$0.009 \\
& PhoenixSR
& 25.94$\pm$0.03 & 0.269$\pm$0.005 & 0.531$\pm$0.005 & 61.7$\pm$0.3 & 0.530$\pm$0.006 \\
\bottomrule
\end{tabular}
}
\end{table*}

Table~\ref{tab:app_seed_robustness} shows that the improvements of PhoenixSR are consistent across independent runs, with small run-to-run variation relative to the observed gains. In particular, the comparison between PhoenixSR with and without DRW verifies that the contribution of reliability weighting is not tied to a single random seed. These results complement the single-run results in the main paper and provide additional evidence that the reported gains are robust to stochastic training variation.

\subsection{Soft Weighting Versus Hard Selection}

We compare DRW with uniform DMD weighting and hard reliability-based sample selection. For hard selection, samples are split according to the same residual-direction consistency used by DRW.

\begin{table*}[t]
\centering

\begin{minipage}[t]{0.49\textwidth}
\vspace{0pt}
\centering
\caption{\textbf{DRW versus uniform weighting and hard sample selection.}}
\label{tab:app_drw}
\vspace{2pt}
\resizebox{\linewidth}{!}{
\begin{tabular}{lcccc}
\toprule
Method
& PSNR$\uparrow$
& MANIQA$\uparrow$
& MUSIQ$\uparrow$
& CLIP-IQA$\uparrow$ \\
\midrule
SFA only
& 26.3124 & 0.5067 & 56.8245 & 0.4538 \\
Uniform DMD w/o DRW
& 26.1587 & 0.5678 & 63.5729 & 0.4963 \\
Low-rel. DMD only
& 26.2072 & 0.5881 & 64.9234 & 0.5173 \\
High-rel. DMD only
& 26.3973 & 0.5526 & 61.9648 & 0.4892 \\
\rowcolor{phoenixgray}
Full PhoenixSR w/ DRW
& 26.4409 & 0.5745 & 64.3102 & 0.5026 \\
\bottomrule
\end{tabular}
}
\end{minipage}
\hfill
\begin{minipage}[t]{0.48\textwidth}
\vspace{0pt}
\centering
\caption{\textbf{Relationship between the DRW score and DMD guidance stability.}
Samples are divided into five equal-sized groups according to the directional consistency score $s_i$.}
\label{tab:app_drw_stability}
\vspace{2pt}
\resizebox{\linewidth}{!}{
\begin{tabular}{lcc}
\toprule
DRW-score group
& Mean $s_i$
& Guidance stability $\uparrow$ \\
\midrule
Lowest 20\%
& -0.604
& 0.074 \\
20--40\%
& -0.195
& 0.091 \\
40--60\%
& 0.206
& 0.173 \\
60--80\%
& 0.579
& 0.418 \\
Highest 20\%
& 0.773
& 0.577 \\
\bottomrule
\end{tabular}
}
\end{minipage}

\end{table*}

Uniform DMD improves perceptual metrics but reduces PSNR relative to SFA alone.
High-reliability-only training preserves fidelity better but provides weaker
perceptual improvement, whereas low-reliability-only training moves further
toward the perceptual regime. DRW retains contributions from both subsets and
softly adjusts their relative strength, obtaining the highest PSNR together
with strong perceptual quality. This also shows that the directional consistency
should be interpreted as a reliability signal for score estimation rather than
a direct image-quality measure.

\subsection{Empirical Validation of DRW Reliability}
\label{app:drw_reliability}

The experiments above show that DRW improves the final SR performance. We further
examine whether the directional consistency score in Eq.~\eqref{eq:app_drw_score}
indeed reflects the reliability of the sample-wise DMD guidance. Recall that the
per-sample DMD guidance $\mathbf{g}_i$ is defined in
Eq.~\eqref{eq:app_dmd_guidance}. Intuitively, if the fake-score model provides
reliable local guidance for a sample, independently sampled diffusion
perturbations should produce more consistent DMD update directions.

\paragraph{Directional consistency versus DMD guidance stability.}
For each validation sample, we first compute its DRW score $s_i$ using one
sampled timestep and Gaussian noise. We then independently sample $K=8$
additional perturbations and compute the corresponding DMD guidance vectors
$\{\mathbf{g}_i^{(1)},\ldots,\mathbf{g}_i^{(K)}\}$.
We measure the guidance stability of sample $i$ by the average pairwise cosine
similarity:
\begin{equation}
\mathrm{Stab}_i
=
\frac{2}{K(K-1)}
\sum_{1\leq a<b\leq K}
\cos\left(
\mathbf{g}_i^{(a)},
\mathbf{g}_i^{(b)}
\right).
\label{eq:app_guidance_stability}
\end{equation}
A larger value indicates that the DMD guidance remains more consistent under
different diffusion perturbations.

We sort samples according to $s_i$ and divide them into five equally sized
groups. Table~\ref{tab:app_drw_stability} reports the average guidance stability
in each group.

As shown in Table~\ref{tab:app_drw_stability}, samples with higher directional consistency exhibit increasingly stable DMD guidance across stochastic perturbations. This observation supports our use of $s_i$ as an empirical reliability proxy: a high DRW score indicates that the online fake-score model produces more consistent local guidance around the generated and paired HR regions. Importantly, this analysis does not interpret $s_i$ as an image-quality measure; instead, it directly relates $s_i$ to the stability of the distributional supervision used to optimize the student.

\paragraph{Comparison with alternative reliability proxies.}
We further examine whether the proposed residual-direction consistency is more informative than simpler alternatives. We compare four sample-wise signals: latent distance between the generated and HR latents, residual distance between $\mathbf{r}_G$ and $\mathbf{r}_{HR}$, cosine similarity between the raw fake-score predictions, and the proposed cosine similarity between denoising residuals. For distance-based measures, we use the negative distance such that a larger value consistently represents higher estimated reliability.

For each proxy, we compute its Spearman rank correlation with the guidance stability defined in Eq.~\eqref{eq:app_guidance_stability}. The results are reported in Table~\ref{tab:app_drw_proxy}.

Table~\ref{tab:app_drw_proxy} shows that the proposed residual-direction
consistency has the strongest correlation with DMD guidance stability among
the tested signals. Compared with latent or residual distances, directional
consistency focuses on whether the fake-score model provides compatible
denoising directions around the generated and HR neighborhoods. Removing the
shared noise component through the residual formulation also makes this
comparison less dominated by the sampled perturbation itself. These results
provide empirical support for using residual-direction consistency, rather
than a generic sample-distance measure, to construct the DRW weights.

\subsection{HDA Score-Model Designs}

Table~\ref{tab:app_hda} compares the score-model configurations used during
development of HDA.

\begin{table*}[t]
\centering

\begin{minipage}[t]{0.45\textwidth}
\vspace{0pt}
\centering
\caption{\textbf{Comparison of sample-wise reliability proxies.}
We report the Spearman rank correlation between each proxy and DMD guidance
stability. A larger positive correlation indicates that the proxy better
reflects guidance stability.}
\label{tab:app_drw_proxy}
\vspace{2pt}
\resizebox{\linewidth}{!}{
\begin{tabular}{lc}
\toprule
Reliability proxy
& Spearman $\rho$ $\uparrow$ \\
\midrule

Latent distance
$\;-\|\mathbf{z}^{G}_0-\mathbf{z}^{HR}_0\|_2$
& 0.33 \\

Residual distance
$\;-\|\mathbf{r}_{G}-\mathbf{r}_{HR}\|_2$
& 0.27 \\

Prediction cosine
$\;\cos(
\boldsymbol{\epsilon}_{\mathrm{fake}}(\mathbf{z}^{G}_t),
\boldsymbol{\epsilon}_{\mathrm{fake}}(\mathbf{z}^{HR}_t)
)$
& 0.41 \\

\rowcolor{phoenixgray}
Residual-direction cosine (DRW)
$\;\cos(\mathbf{r}_{G},\mathbf{r}_{HR})$
& \textbf{0.59} \\

\bottomrule
\end{tabular}
}
\end{minipage}
\hfill
\begin{minipage}[t]{0.53\textwidth}
\vspace{0pt}
\centering
\caption{\textbf{Comparison of HDA score-model designs.}
We compare different configurations of the real and fake score models, including full fine-tuning, REPA adaptation, LoRA adaptation, and SFA.}
\label{tab:app_hda}
\vspace{2pt}
\resizebox{\linewidth}{!}{
\begin{tabular}{lcccc}
\toprule
Method
& PSNR$\uparrow$
& MANIQA$\uparrow$
& MUSIQ$\uparrow$
& CLIP-IQA$\uparrow$ \\
\midrule

Frozen Real + Full Fake w/o SFA
& 21.2561 & 0.4023 & 39.6617 & 0.3569 \\

Frozen Real + Full Fake + SFA
& 25.9778 & 0.5422 & 63.9735 & 0.4931 \\

Frozen Real + REPA Fake + SFA
& 26.3012 & 0.5787 & 64.1157 & 0.4898 \\

\rowcolor{phoenixgray}
LoRA Real + REPA Fake + SFA
& 26.4409 & 0.5745 & 64.3102 & 0.5026 \\

\bottomrule
\end{tabular}
}
\end{minipage}

\end{table*}

DRW is kept enabled for all variants in this comparison so that only the score-model designs within HDA are varied. Without SFA, direct distribution matching produces severe degradation, showing that paired reconstruction constraints are critical for fidelity-sensitive SR. Adding SFA restores stable reconstruction and substantially improves perceptual quality. Representation regularization further improves PSNR and MANIQA, while the final LoRA-adapted real-score configuration achieves the strongest PSNR and CLIP-IQA with competitive perceptual scores. Together, these results support the use of fidelity anchoring, representation-regularized fake-score tracking, and lightweight real-score adaptation.

\subsection{Effect of the Training Schedule}

Removing timestep truncation causes the largest PSNR degradation, supporting
the use of low-noise diffusion states for fidelity-sensitive SR. Removing
either the noise-shift strategy or warm-up produces smaller but consistent
performance drops. The full schedule gives the strongest overall result among
the tested variants.

\subsection{Comparison with Conventional Distillation}

We additionally compare PhoenixSR with conventional feature- and output-level
knowledge distillation. Feature KD matches intermediate diffusion and DFSR
representations using lightweight adapters, while output KD directly supervises
the DFSR student with sampled diffusion-SR outputs.

\paragraph{Scope of the distribution-matching baselines.}
We use \emph{DMD-only} to denote the unanchored score-difference ablation without paired SR fidelity losses or teacher-output regression, referred to as \emph{vanilla DMD} in the main text and Fig.~\ref{fig:vanilla_dmd_failure}. This ablation isolates the role of fidelity anchoring in our SR setting; it is not a reproduction of the complete DMD or DMD2 training recipe~\citep{yin2024one,yin2024improved}. The \emph{DMD2 + SFA} baseline in Table~\ref{tab:transfer_comparison} is a DMD2-based adaptation to native DFSR students with paired SR supervision. Its results concern this controlled transfer setting rather than the original diffusion-derived generators.

\begin{table*}[t]
\centering

\begin{minipage}[t]{0.49\textwidth}
\vspace{0pt}
\centering
\caption{\textbf{Ablation of the training schedule.}}
\label{tab:app_schedule}
\vspace{2pt}
\resizebox{\linewidth}{!}{
\begin{tabular}{lcccc}
\toprule
Method
& PSNR$\uparrow$
& MANIQA$\uparrow$
& MUSIQ$\uparrow$
& CLIP-IQA$\uparrow$ \\
\midrule

w/o timestep truncation
& 25.6564 & 0.5439 & 61.6731 & 0.4733 \\

w/o noise-shift curriculum
& 26.3797 & 0.5503 & 62.7315 & 0.4977 \\

w/o warm-up
& 26.4011 & 0.5637 & 64.1149 & 0.4893 \\

\rowcolor{phoenixgray}
Full schedule
& 26.4409 & 0.5745 & 64.3102 & 0.5026 \\

\bottomrule
\end{tabular}
}
\end{minipage}
\hfill
\begin{minipage}[t]{0.49\textwidth}
\vspace{0pt}
\centering
\caption{\textbf{Comparison with conventional distillation strategies.}}
\label{tab:app_conventional_kd}
\vspace{2pt}
\resizebox{\linewidth}{!}{
\begin{tabular}{lcccc}
\toprule
Method
& PSNR$\uparrow$
& MANIQA$\uparrow$
& MUSIQ$\uparrow$
& CLIP-IQA$\uparrow$ \\
\midrule

SFA only
& 26.3124 & 0.5067 & 56.8245 & 0.4538 \\

Feature KD
& 26.1638 & 0.5132 & 53.7781 & 0.4569 \\

Output KD
& 25.9754 & 0.4983 & 56.7369 & 0.4691 \\

DMD-only
& 21.2561 & 0.4023 & 39.6617 & 0.3569 \\

\rowcolor{phoenixgray}
PhoenixSR
& 26.4409 & 0.5745 & 64.3102 & 0.5026 \\

\bottomrule
\end{tabular}
}
\end{minipage}

\end{table*}

Feature KD provides only limited and inconsistent gains, reflecting the weak
correspondence between diffusion and feed-forward SR representations. Output KD
slightly improves CLIP-IQA but degrades PSNR, MANIQA, and MUSIQ relative to SFA,
indicating that direct imitation of sampled diffusion outputs does not provide
a consistently favorable fidelity--perception trade-off. DMD-only shows
severe fidelity degradation in this unanchored SR setting. PhoenixSR achieves
the strongest overall balance, supporting distribution-level transfer with
fidelity and reliability control.

\section{Additional Results}
\label{app:additional_results}

\subsection{Additional Quantitative Discussion}

\paragraph{RealSR.}
For medium and large students, PhoenixSR preserves or improves PSNR while consistently strengthening perceptual quality. SwinIR-L, HAT, Real-ESRGAN, and SwinIR-M improve PSNR by $0.44$, $0.22$, $0.26$, and $0.45$ dB, respectively. Compact SwinIR-S and SeeMoRe remain within $0.2$ dB of their original PSNR while obtaining substantially stronger no-reference quality scores.

\paragraph{DRealSR.}
SwinIR-L and SwinIR-M gain $0.65$ and $0.54$ dB in PSNR, while all medium/large backbones improve across the reported no-reference metrics. The perceptual gain is especially strong for compact students: SeeMoRe improves MUSIQ by $16.27$, LIQE-Mix by $1.78$, ARNIQA by $0.255$, and TOPIQ by $0.188$.

\paragraph{DIV2K-Val.}
PhoenixSR produces a stronger distortion--perception trade-off on DIV2K-Val. SwinIR-L and SwinIR-M improve both PSNR and perceptual quality. SwinIR-S and SeeMoRe exhibit PSNR changes of only $-0.14$ and $-0.19$ dB while improving MUSIQ by $17.73$ and $23.33$, respectively. SeeMoRe also improves LPIPS by $0.179$.

These results further indicate that diffusion-prior transfer is particularly beneficial for compact models whose limited capacity otherwise restricts realistic high-frequency reconstruction.

\subsection{Additional Qualitative Comparisons}

We provide additional qualitative comparisons on RealSR and DRealSR to complement the quantitative results in the main paper. The examples cover different scene contents and local structures, and compare the original DFSR backbones with their PhoenixSR-distilled counterparts. Across different student architectures, PhoenixSR generally recovers clearer high-frequency details and more natural textures while preserving the underlying image structures.

\begin{figure*}[t]
    \centering
    \includegraphics[width=\textwidth]{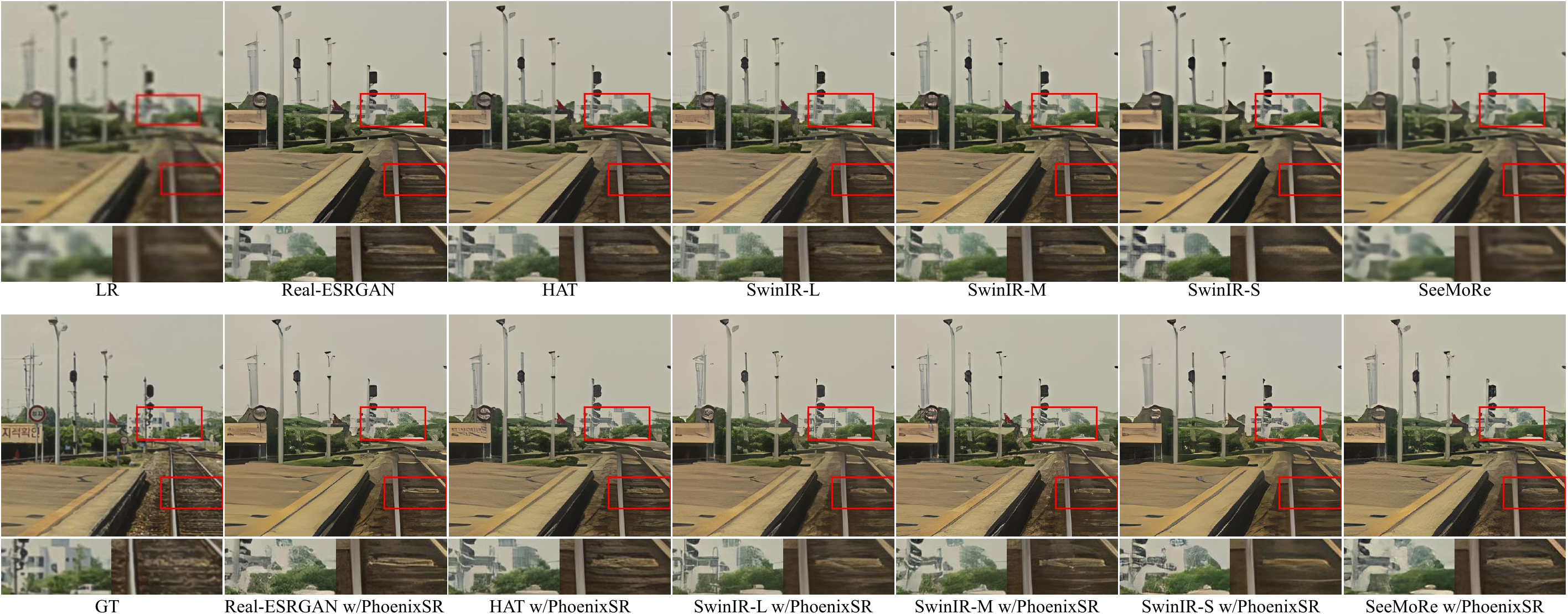}
    \caption{
    \textbf{Additional qualitative comparisons on RealSR.}
    PhoenixSR improves fine structures and local texture reconstruction across different feed-forward SR backbones while preserving the underlying scene content and geometry. Best viewed by zooming in.
    }
    \label{fig:app_visual_realsr}
\end{figure*}

\paragraph{RealSR.}
Figure~\ref{fig:app_visual_realsr} presents additional qualitative comparisons on the RealSR dataset. Compared with the original feed-forward SR models, PhoenixSR improves the restoration of fine structures and local textures across different student backbones. In particular, distant structures and rail-like textures become more distinguishable after distillation, while the overall geometry and scene content remain consistent with the original reconstruction. These visual improvements are consistent with the quantitative results reported on RealSR.

\begin{figure*}[t]
    \centering
    \includegraphics[width=\textwidth]{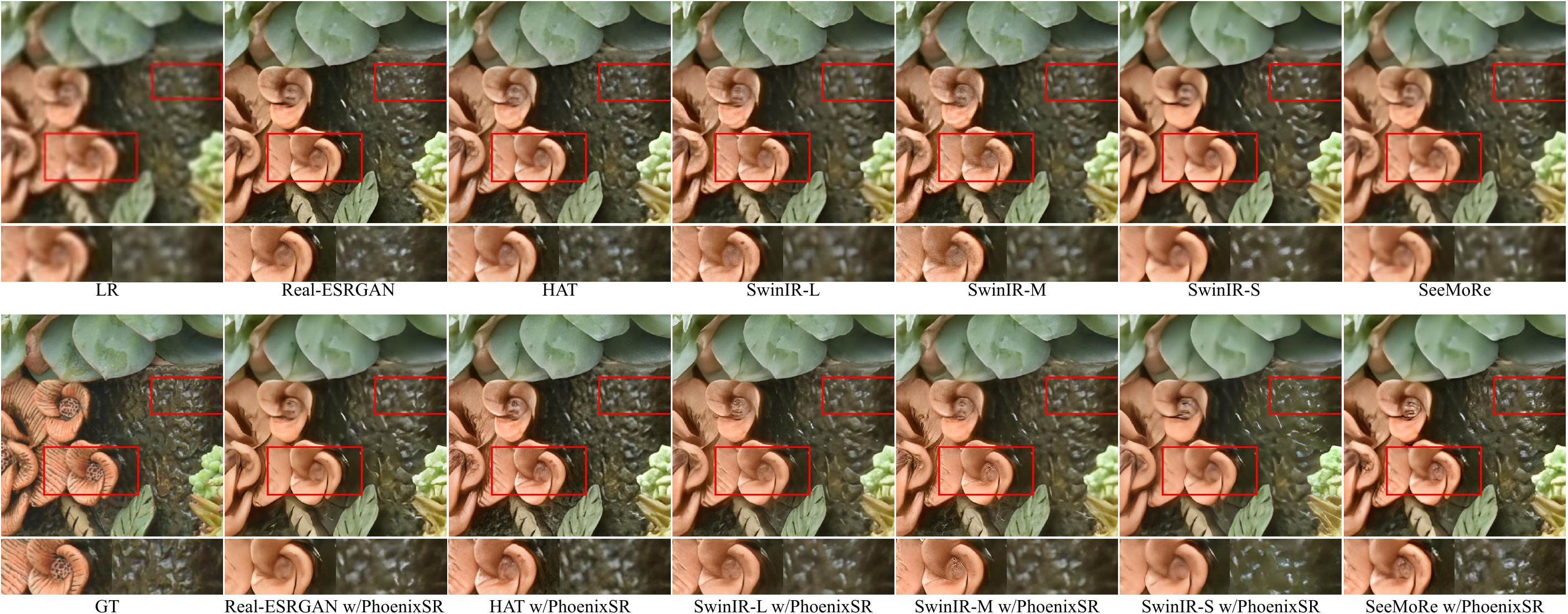}
    \caption{
    \textbf{Additional qualitative comparisons on DRealSR.}
    PhoenixSR recovers clearer local structures and more natural texture details across different feed-forward SR backbones while largely preserving reconstruction fidelity. Best viewed by zooming in.
    }
    \label{fig:app_visual_drealsr}
\end{figure*}

\paragraph{DRealSR.}
Figure~\ref{fig:app_visual_drealsr} shows additional qualitative results on DRealSR. PhoenixSR produces clearer local structures and more recognizable texture details, particularly in regions that are easily over-smoothed by the original DFSR models. The improvements are observed across different student architectures, further demonstrating that the proposed diffusion-prior transfer is not tied to a specific feed-forward SR backbone.

Overall, these qualitative results are consistent with the quantitative observations across architectures. Improvements can be observed for CNN, Transformer, and lightweight students, providing additional evidence that PhoenixSR can transfer useful diffusion priors to different native feed-forward SR architectures without modifying their inference pipelines.

\section{User Study Details}
\label{app:userstudy}

\begin{figure}[t]
    \centering
    \includegraphics[width=\linewidth]{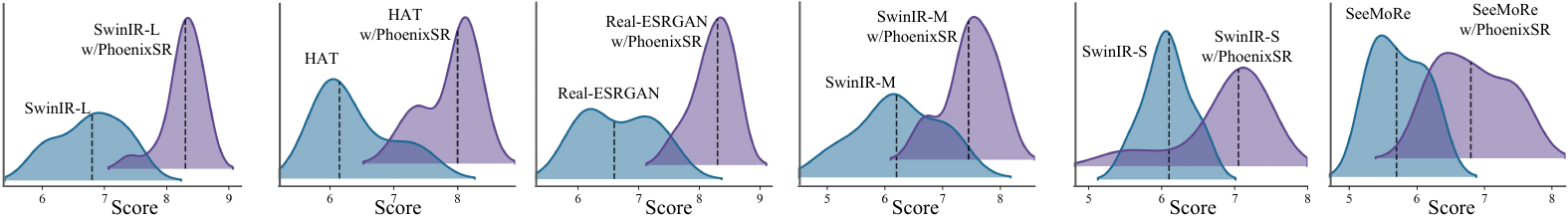}
    \caption{\textbf{User study} across six DFSR backbones.}
    \label{fig:user_study}
\end{figure}

\paragraph{Study protocol.}
We conduct a subjective study with 20 adult participants using 20 representative source images, including 7 images from RealSR, 7 from DRealSR, and 6 from DIV2K-Val. For each source image and DFSR backbone, the outputs before and after PhoenixSR distillation are presented in randomized left--right order to reduce positional bias. Participants are not informed which output is produced by PhoenixSR.

Participants score the overall visual quality of each result on a scale from $0$ to $10$, considering texture naturalness, detail sharpness, and artifact suppression. Each participant evaluates 120 paired comparisons, corresponding to 20 source images and six DFSR backbones. This results in 2,400 paired evaluations and 4,800 individual output ratings in total. All participants participate voluntarily after being informed of the purpose and procedure of the study, and no personally identifiable information is collected.

\paragraph{Image-level statistical analysis.}
For each source image and backbone, we first average the subjective scores from the 20 participants, yielding one image-level score for the original model and one for PhoenixSR. We report the mean score across the 20 source images together with a 95\% confidence interval estimated by bootstrap resampling over images with 10,000 repetitions. Statistical significance is evaluated using a two-sided Wilcoxon signed-rank test on the paired image-level scores.

\begin{table}[H]
\centering
\caption{\textbf{Detailed user-study statistics.}
Scores are reported as image-level mean scores with 95\% bootstrap confidence intervals. $\Delta$ denotes PhoenixSR minus the original backbone.}
\label{tab:app_userstudy_stats}
\resizebox{\linewidth}{!}{
\begin{tabular}{lccc}
\toprule
Backbone
& Original
& PhoenixSR
& $\Delta$ (95\% CI) \\
\midrule
Real-ESRGAN
& 6.38 [6.14, 6.64]
& 7.89 [7.71, 8.06]
& +1.51 [1.36, 1.66] \\
HAT
& 6.78 [6.56, 7.00]
& 8.29 [8.15, 8.40]
& +1.51 [1.39, 1.63] \\
SwinIR-L
& 6.67 [6.43, 6.92]
& 8.24 [8.10, 8.37]
& +1.57 [1.44, 1.70] \\
SwinIR-M
& 6.22 [5.92, 6.51]
& 7.50 [7.31, 7.67]
& +1.28 [1.14, 1.42] \\
SwinIR-S
& 6.08 [5.95, 6.21]
& 6.87 [6.60, 7.11]
& +0.79 [0.63, 0.92] \\
SeeMoRe
& 5.74 [5.58, 5.89]
& 6.81 [6.60, 7.04]
& +1.07 [0.87, 1.31] \\
\midrule
Overall
& 6.31 [6.10, 6.52]
& 7.60 [7.43, 7.75]
& +1.29 [1.19, 1.38] \\
\bottomrule
\end{tabular}
}
\end{table}

Across the 20 source images, PhoenixSR obtains a higher overall subjective score than the corresponding original DFSR outputs (7.60 versus 6.31), with a mean paired improvement of 1.29 points (95\% CI: [1.19, 1.38]). The paired image-level difference is statistically significant under a two-sided Wilcoxon signed-rank test ($p<0.001$). Figure~\ref{fig:user_study} and Table~\ref{tab:app_userstudy_stats} further show that the subjective improvement is consistently observed across all six evaluated backbones.

\paragraph{Participant-level robustness analysis.}
We additionally average each participant's scores across all evaluated source images and backbones, yielding one paired Original/PhoenixSR score for each of the 20 participants. At the participant level, the mean subjective score increases from 6.31 for the original DFSR outputs to 7.60 for PhoenixSR, corresponding to an average improvement of 1.29 points. Bootstrap resampling over participants with 10,000 repetitions gives a 95\% confidence interval of [1.05, 1.47] for the paired improvement. The difference remains statistically significant under a two-sided Wilcoxon signed-rank test ($p<0.001$), and 18 of the 20 participants assign a higher average score to PhoenixSR. This complementary analysis indicates that the observed preference is consistent across participants rather than being driven only by a small subset of source images.

The subjective study is used only as complementary evidence of perceptual quality and does not replace the objective fidelity and perceptual evaluations reported in the main paper.

\end{document}